%% file: main.tex
\documentclass[10pt,twocolumn,letterpaper]{article}

\usepackage{iccv}              
\usepackage{algorithm}
\usepackage{algorithmic}
\usepackage{multirow}
\usepackage{bbding}
\usepackage{xcolor}
\usepackage[none]{hyphenat}
\usepackage[accsupp]{axessibility}


\definecolor{iccvblue}{rgb}{0.21,0.49,0.74}
\usepackage[pagebackref,breaklinks,colorlinks,allcolors=iccvblue]{hyperref}

\def\paperID{13538} 
\def\confName{ICCV}
\def\confYear{2025}

\title{Make Me Happier: Evoking Emotions Through Image Diffusion Models}

\author{
Qing Lin\textsuperscript{1} \quad
Jingfeng Zhang\textsuperscript{2,3} \quad
Yew-Soon Ong\textsuperscript{1} \quad
Mengmi Zhang\textsuperscript{1,*} \\
\textsuperscript{1}Nanyang Technological University, Singapore \\
\textsuperscript{2}The University of Auckland, New Zealand \\
\textsuperscript{3}RIKEN AIP, Tokyo, Japan \\
\textsuperscript{*} Address correspondence to mengmi.zhang@ntu.edu.sg \\
}

\begin{document}

\twocolumn[{%
\renewcommand\twocolumn[1][]{#1}%
\maketitle
\begin{center}
    \centering
    \includegraphics[width=1\linewidth]{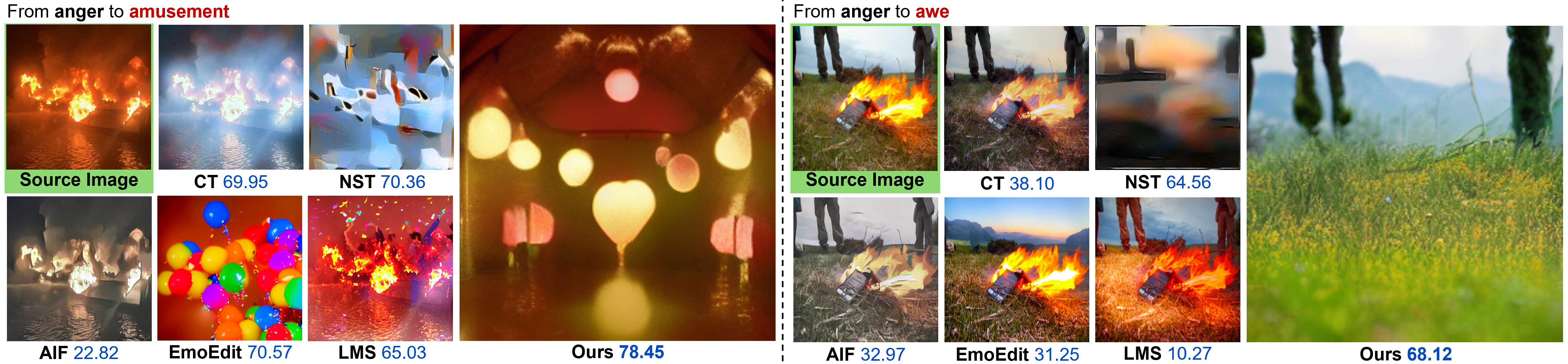}
    \vspace{-4mm}
    \captionof{figure}{\textbf{The generated images evoke a sense of happiness in viewers, contrasting with the negative emotions elicited by the source images.} 
    Given a source image that triggers negative emotions (framed in green), our method (Ours) synthesizes a new image that elicits the given positive target emotions (in red), while maintaining the essential elements and structures of the scene. 
    For instance, our method replaces the anger-inducing flames in the source image with cute-shaped lamps to evoke the target emotion of amusement. While in an outdoor setting, the raging fire is substituted with a tranquil, lush meadow to inspire a sense of awe. 
    For comparisons, we include other competitive methods. The blue number below each image represents its CAM-based ESMI score, with higher values being better. 
    }%
    \vspace{-1mm}
    \label{fig:fig1_teaser}
\end{center}
}]

\sloppy

\begin{abstract}

Despite the rapid progress in image generation, emotional image editing remains under-explored. The semantics, context, and structure of an image can evoke emotional responses, making emotional image editing techniques valuable for various real-world applications, including treatment of psychological disorders, commercialization of products, and artistic design. First, we present a novel challenge of emotion-evoked image generation, aiming to synthesize images that evoke target emotions while retaining the semantics and structures of the original scenes. To address this challenge, we propose a diffusion model capable of effectively understanding and editing source images to convey desired emotions and sentiments. Moreover, due to the lack of emotion editing datasets, we provide a unique dataset consisting of 340,000 pairs of images and their emotion annotations. Furthermore, we conduct human psychophysics experiments and introduce a new evaluation metric to systematically benchmark all the methods. Experimental results demonstrate that our method surpasses all competitive baselines. Our diffusion model is capable of identifying emotional cues from original images, editing images that elicit desired emotions, and meanwhile, preserving the semantic structure of the original images. All code, model, and dataset are available at \href{https://github.com/ZhangLab-DeepNeuroCogLab/EmoEditor}{GitHub}. 

\end{abstract}

\vspace{-4mm}
\section{Introduction}
\label{sec:intro}

\noindent \textit{``I am feeling down. Bring some excitement to my room."}

What we perceive not only entails useful visual information but also evokes profound emotional reactions. Incorporating visual cues that elicit emotions into images holds practical significance in various domains. 
In commercialization, it enhances brand impact and captures consumer attention. In augmented or virtual reality, it supports therapeutic interventions for conditions such as autism and schizophrenia. 
Despite recent advances in image generation \cite{gal2022stylegan,wang2023imagen,yu2018generative}, emotion conditioning remains largely underexplored. We introduce the novel task of emotion-evoked image generation: given a source image and a target emotion, the goal is to generate a new image that evokes the specified emotion while preserving the semantics and structure of the original scene. 
This task is challenging, requiring both recognition of subtle emotional cues in the source image and the ability to modify them meaningfully. For example, as shown in Figure \ref{fig:fig1_teaser}, clusters of fire flames on the water intensify the emotion of anger. In the generated image, they are replaced with lamps designed in cute shapes, transforming the overall emotional response to amusement.

\begin{figure}
    \centering
    \includegraphics[width=0.99\linewidth]{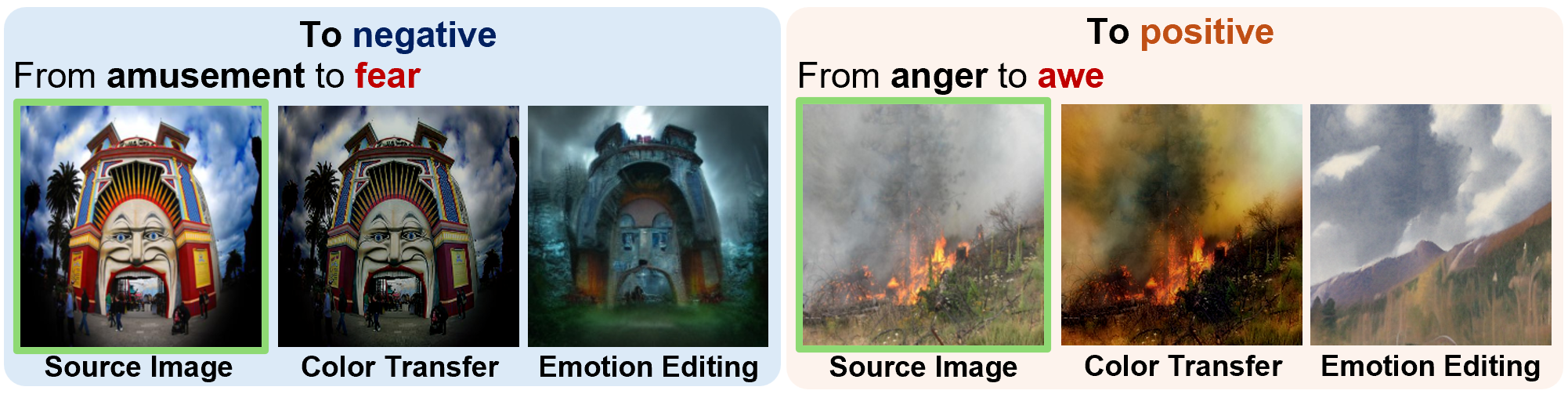}
    \vspace{-2.5mm}
    \caption{
    \textbf{Emotions are influenced by both global and local cues from the source images.} 
    We present examples of transitions from positive to negative emotions (blue shading) and from negative to positive emotions (orange shading). 
    In each example, the target emotions are in red, and source images are framed in green.  
    }
    \vspace{-4mm}
    \label{fig:fig2_intro}
\end{figure}

Studies have shown that both global and local factors influence emotional responses in images, such as overall color tones, local facial expressions, and emotion-related objects like graveyards (sadness) or balloons (amusement) \cite{yang2023emoset,yang2018visual}. 
As shown in Figure \ref{fig:fig2_intro}, inserting fearful elements, such as fire, into a peaceful scene can evoke negative emotions, as they trigger instinctive responses to threats. However, some elements, such as the sky and plants, remain emotionally neutral. Given that humans are generally more sensitive to negative than positive stimuli \cite{baumeister2001bad,ito1998negative}, simply adjusting global attributes like brightness may not be enough to alter negative emotions. 
Traditional methods such as color transfer and style transfer \cite{pitie2007automated,gatys2015neural,weng2023affective} attempt to modify emotional tone through global appearance changes. However, they often fail to elicit desired emotions, as they cannot identify and edit the specific local regions responsible for emotional impact. In Figure \ref{fig:fig2_intro}, while darkening the scene may enhance fear (left panel), brightening a scene of a burning hillside does not effectively convey awe (right panel). 
In contrast, transforming anger into awe requires editing key emotional cues, for example, removing the fire and adding vibrant clouds to create a tranquil atmosphere. Emotion-evoked image generation thus requires not just global color or style adjustments, but also semantic understanding and purposeful local modification.

Drawing inspiration from these emotion-associated factors, we propose our model, EmoEditor, for emotion-evoked image generation. We develop EmoEditor based on state-of-the-art stable diffusion models and introduce three technical novelties. (1) EmoEditor employs a novel dual-branch architecture in the reverse diffusion process that integrates emotion-conditioned global context and local emotional cues from the source images to ensure coherence with the original content while reflecting the desired emotional outcome. (2) During training, the network's behaviors are guided by aligning model creativity with human expectations in the neuro-symbolic space, implicitly learning the mapping from target emotions to human-annotated text instructions. (3) During inference, iterative emotion discrimination mechanisms are employed to select emotionally coherent images autonomously.

With the rapid advancements in diffusion models, various effective image editing techniques have emerged \cite{kawar2023imagic,brooks2023instructpix2pix,ruiz2023dreambooth}. However, these methods are not specifically tailored for understanding and identifying emotion-associated elements within images. To address this gap, we propose a competitive baseline called Large Model Series (LMS), concatenating multiple large language and vision models. This baseline first understands source images through image captioning, then generates emotion-prompted instructions using language models, and finally edits images prompted by these instructions with diffusion models. Upon comparison with existing methods through a series of human psychophysics experiments and a new quantitative evaluation, our end-to-end trained EmoEditor achieves outstanding image generation results. It effectively preserves the structural coherence and semantic consistency with the source images while eliciting the target emotion of human viewers. 
To obtain training data for this task, we establish a dataset curation pipeline and introduce the first large-scale EmoPair dataset, comprising approximately 340,000 pairs of images annotated with emotion-conditioned editing instructions. Unlike existing datasets on visual emotion analysis \cite{mikels2005emotional,peng2015emotion6,yang2023emoset}, which assign emotion labels to individual images for classification, our dataset provides paired images annotated with source and target emotion labels.

The main contributions of this paper are highlighted below: (1) We study the problem of emotion-evoked image generation, which focuses on modifying visual content to trigger intended emotional responses while preserving scene structures and contextual coherence. 
To benchmark all methods, we establish a comprehensive framework for evaluating model performance, incorporating human psychophysics experiments and introducing a new metric. 
(2) We propose EmoEditor, an emotion-evoked diffusion model, which employs a novel two-branch architecture to integrate global context with local semantic regions of source images that evoke emotional responses. 
(3) EmoEditor learns emotional cues during end-to-end training, eliminating the need for additional emotion reference images. During inference, EmoEditor generates emotion-evoked images without hand-crafted emotion-conditioned text instructions, while preserving context and structural coherence with the source image. Its performance is superior among all the competitive methods. 
(4) We curate the large-scale EmoPair dataset, containing 340,000 image pairs with emotion annotations.

\section{Related Work}
\subsection{Visual Emotion Analysis}
Visual Emotion Analysis (VEA) seeks to understand and predict human emotional responses from visual data. In psychology, Categorical Emotion States (CES) underpin classification models such as the 8-category Mikels model \cite{mikels2005emotional}. Early work relied on handcrafted features like color \cite{machajdik2010affective}, while recent methods leverage deep learning for more accurate emotion prediction \cite{you2015robust,yang2018visual}. However, generating images to evoke specific emotions remains underexplored. To address this, we introduce EmoEditor, a two-branch diffusion model that jointly learns emotion recognition and image generation. 

\textbf{Existing VEA datasets} \cite{peng2015emotion6,you2016building} focus on emotion classification, including EmoSet \cite{yang2023emoset}, which has 120k images with emotion labels and attributes. However, they lack image pairs with annotated emotions and text editing instructions essential for emotion-evoked image generation. Thus, we introduce the EmoPair dataset, comprising 340k image pairs with differing source and target emotions while ensuring consistent scene semantics and structure.

\subsection{Image Editing}
\textbf{Diffusion-based Image Editing.} Earlier studies used GANs \cite{goodfellow2014GAN} and CLIP models \cite{radford2021CLIP} for image manipulation based on detailed text descriptions \cite{abdal2022clip2stylegan}. 
Diffusion models have advanced, generating high-quality images from various prompts \cite{hertz2022prompt, ruiz2023dreambooth}. Methods like SDEdit \cite{meng2021sdedit} use stochastic differential equations for editing. Ip2p \cite{brooks2023instructpix2pix} allows image editing based on human instructions. PnP \cite{tumanyan2023plug} achieves fine-grained control of generated structures by manipulating spatial features and self-attention. CtrlNet \cite{zhang2023ControlNet} adds conditional control for better precision. Despite these advances, current diffusion models often overlook emotional cues needed for effective emotion editing.

\noindent \textbf{Emotion Editing.} Color and style transfer methods like PDF-Transfer \cite{pitie2007automated}, Neural Style Transfer (NST) \cite{gatys2015neural}, and CLIP-Styler (Csty) \cite{kwon2022clipstyler} manipulate images using reference images but struggle with local adjustments, limiting their versatility in generating emotion-evoking images. AIF \cite{weng2023affective} reflects emotions from text to images by incorporating affective content into the image generation process. It relies heavily on detailed textual prompts and cannot edit image regions locally. 
EmoEdit \cite{yang2025emoedit} maps input images and target emotions into an embedding guided by a set of fixed learned queries, restricting emotional cues and diversity in generation to the representational capacity of these queries. In contrast, our EmoEditor uses end-to-end training to implicitly understand and edit emotional cues locally and globally without relying on external references.

\section{Method}
\label{sec:method}
We formulate emotion-evoked image generation as a supervised learning task. First, we curate the EmoPair dataset with emotion-labeled image pairs and editing instructions. 
Next, EmoEditor is trained on this dataset. 
During inference, it takes source images and target emotions, producing generated images \textit{without} the need for text instructions or emotion reference images.

\subsection{Curating Our EmoPair Dataset}
\label{sec:EmoPair}
\begin{figure*}[ht]
	\centering
	\includegraphics[width=0.89\linewidth]{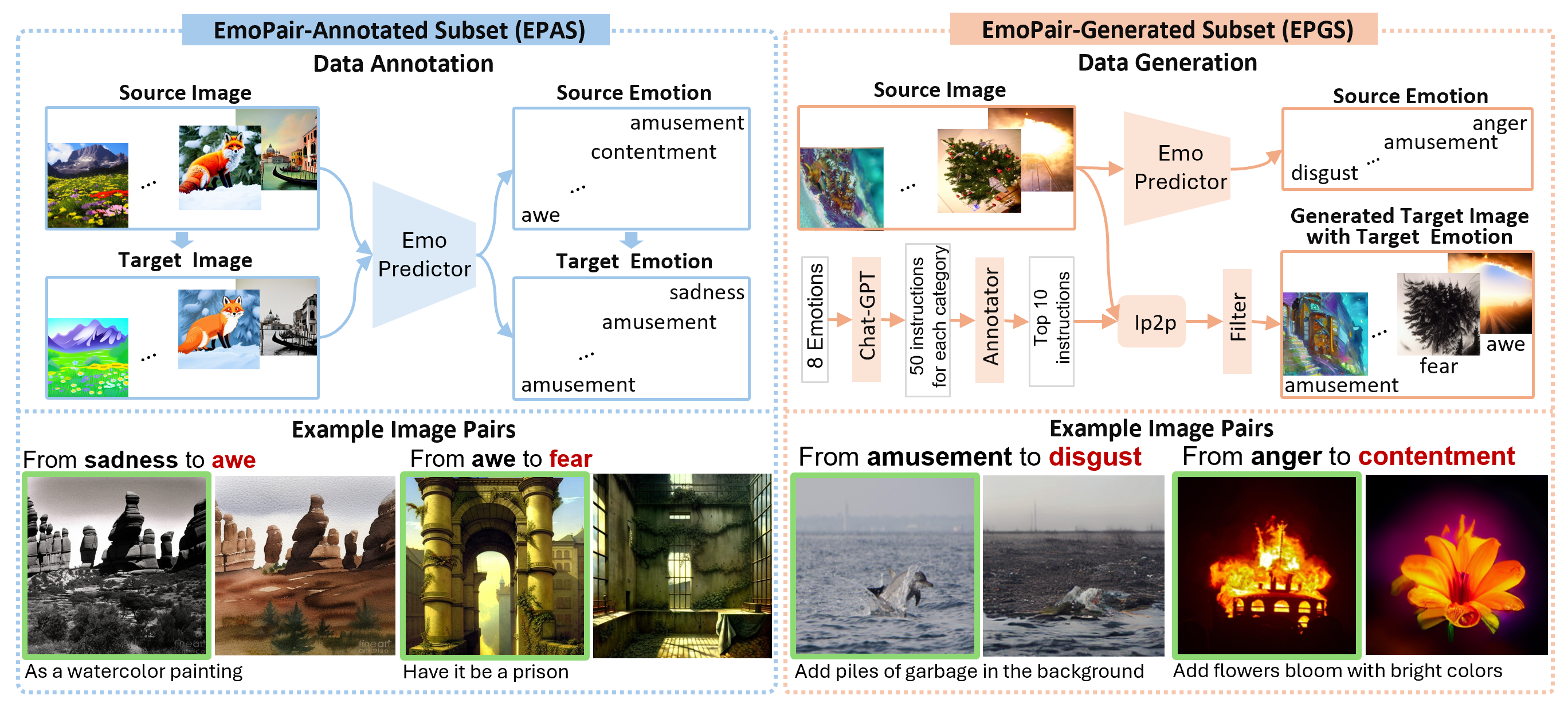}
        \vspace{-2.5mm}
	\caption{\textbf{Pipeline for Curating Our EmoPair Dataset.} The dataset comprises two subsets: EmoPair-Annotated Subset (EPAS, left blue box) and EmoPair-Generated Subset (EPGS, right orange box). Each subset includes schematics depicting the creation, selection, and labeling of image pairs in the upper quadrants, with two example pairs in the lower quadrants. Each example pair comprises a source image (framed in green) and a target image. The classified source and target emotion labels (highlighted in red) and target-emotion-driven text instructions for image editing are provided. 
    Note that these instructions are used during model training to guide the learning of rich neural-symbolic embeddings, but are not used during model inference.
    }
    \label{fig:fig4_dataset}
    \vspace{-3mm}
\end{figure*}

Given the unavailability of a paired image dataset for emotion editing, we curate EmoPair. It consists of two subsets: EmoPair-Annotated Subset (EPAS), containing 331,595 image pairs from Ip2p \cite{brooks2023instructpix2pix} annotated with emotion labels; 
and EmoPair-Generated Subset (EPGS), consisting of 6,949 pairs generated based on text instructions given target emotions. All images from EmoPair are of size 224$\times$224. Figure \ref{fig:fig4_dataset} illustrates the pipeline for curating our EmoPair dataset. See \textbf{Sec S1} for more details. 

\noindent \textbf{Emotion Predictor $\mathcal{P}$.} 
To construct our dataset, we use the emotion predictor $\mathcal{P}$ proposed in \cite{yang2023emoset}, which was trained on the largest VEA dataset, EmoSet. 
All images are classified into 8 emotion categories, encompassing four positive (amusement, awe, contentment, excitement) and four negative (anger, disgust, fear, sadness) emotions. 

\noindent \textbf{EmoPair-Annotated Subset (EPAS).} The original Ip2p dataset lacks emotion labels, consisting only of image pairs with text instructions for editing. To address this, we use $\mathcal{P}$ to classify the source and target images from the Ip2p dataset into eight distinct emotions. We augment the Ip2p dataset with emotion labels, resulting in 331,595 image pairs. Notably, the text instructions with these image pairs serve as human expectations to guide EmoEditor's training but are not utilized during inference.

\noindent \textbf{EmoPair-Generated Subset (EPGS).} To address the lack of emotional cues in the Ip2p dataset, we create the EPGS subset using emotionally rich images from EmoSet \cite{yang2023emoset}. However, EmoSet lacks paired images with consistent scenes and target emotions, and it does not include text editing instructions. Single-word prompts like ``[Target Emotion]" prove ineffective in existing text-to-image models due to their limited emotional reasoning. Single-emotion terms lack specificity, resulting in varied interpretations depending on the context of source images. For instance, enhancing an image for excitement can vary significantly based on context, such as having a happy smile on a person or adding fireworks to the sky for a landscape. Thus, to generate accurate image pairs, we develop 50 general instructions for transitioning to desired emotions of 8 categories using GPT-3 \cite{brown2020GPT3}. Human annotators rank these instructions, keeping the top ten per emotion. The pre-trained Ip2p model then uses these instructions to manipulate EmoSet source images, resulting in 6,949 high-quality image pairs after quality control measures. To ensure the quality of the generated images, we have human annotators providing corrections for some results (see \textbf{Sec S1}).

\subsection{Our Diffusion Model - EmoEditor}
\label{sec:EmoEditor}
\begin{figure}[t]
	\centering
	\includegraphics[width=0.99\linewidth]{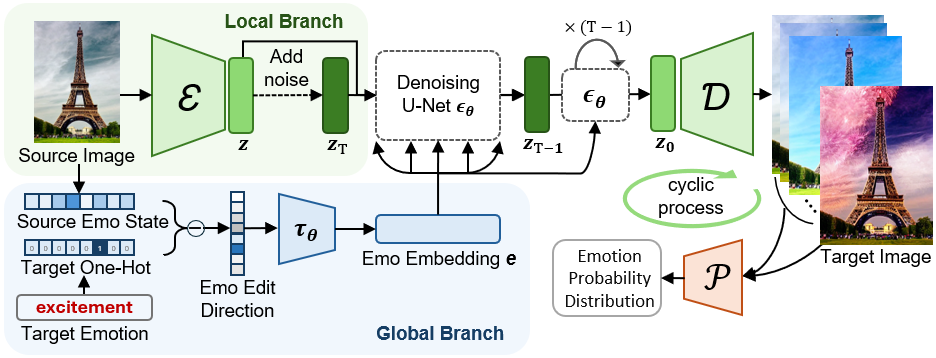}
        \vspace{-2.5mm}
	\caption{\textbf{Architecture of Our Proposed EmoEditor.} EmoEditor is an image diffusion model with local (shaded in green) and global (shaded in blue) branches. The pre-trained VAE's encoder $\mathcal{E}$ and decoder $\mathcal{D}$ are fixed during training and inference. Exclusively employed during inference, the fixed emotion predictor $\mathcal{P}$ predicts emotions on generated images for iterative emotion inference. 
    }
    \vspace{-4mm}
    \label{fig:fig3_model}
\end{figure}

Our EmoEditor, in Figure \ref{fig:fig3_model}, takes a source image $I_s$ and a target emotion $y$ to generate an image evoking the desired emotion. It employs the latent diffusion model (LDM) \cite{rombach2022high} with three key novelties.

\noindent \textbf{Integration of Global and Local Emotion Cues.} 
In the global branch of our EmoEditor, we convert the target emotion $y$ into a binary one-hot vector $e_{oh}$, where the corresponding entry to $y$ in $e_{oh}$ is set to 1 and the rest are set to 0. This simplifies model design, enhancing comprehension and generation of complex emotional images, as one-hot encoding can capture intricate emotion combinations through a multi-label representation. 
We use the pre-trained emotion predictor $\mathcal{P}$ to predict the source image's emotional state $e_s$, representing the probability distribution of emotions for each category. EmoEditor avoids using emotion labels for $I_s$ for two reasons: (1) Obtaining emotion labels for source images is often impractical in real-world scenarios. (2) Source images can be complex, conveying multiple emotions simultaneously. Utilizing a stereotypical source emotion as input potentially introduces biases and limits the model's ability to generate diverse outputs.

Next, we compute the emotion editing direction $e_{dir}$ using the target emotion code $e_{oh}$ and the source emotional state $e_s$: $e_{dir} = e_{oh} - e_s$. 
This allows the model to adaptively adjust the editing direction based on the initial state of different source images. To map the emotion editing direction into the latent space, we introduce an emotion encoder $\tau_{\theta}$, trained from scratch to learn neural-symbolic embeddings. It first encodes $e_{dir}$ into an emotion embedding $e$ through fully connected layers: $e = \tau_{\theta}(e_{dir})$ (See \textbf{Sec S2.1} for details). Later on, we introduce novel losses to regularize the behaviors of $\tau_{\theta}$. Notably, we intentionally avoid using explicit text editing instructions as conditional inputs to the model due to the varied interpretations of target emotions, where multiple solutions can elicit the same emotions. Including explicit text instructions may hinder the model's creativity, restricting its capacity to produce diverse sets of images conveying the same emotions.

In the local branch of our EmoEditor, we introduce an image encoder $\mathcal{E}$ to extract visual features of the source image, acquiring $z$ for $I_s$: $z = \mathcal{E}(I_s)$. Similar to the LDM \cite{rombach2022high}, $\mathcal{E}$ is based on a pre-trained VAE \cite{kingma2013auto}. The forward diffusion process gradually adds noise to $z$, generating a noisy latent $z_t$, where the noise level increases over $t$ time steps ($t\in T$). Specifically, we expressed $z_t = \sqrt{\alpha_t} \cdot z_{t-1} + \sqrt{1 - \alpha_t} \cdot \epsilon$, where $\alpha_t$ is the diffusion coefficient controlling the rate of noise increase, and $\epsilon$ is a random noise sampled from a normal distribution.

In the reverse diffusion process, the denoising network $\epsilon_{\theta}$ predicts noise added to the noisy latent $z_t$ using cross-attention with the target emotion condition $e$ and source image condition $z$. The features in $z$ and $z_t$ enhance understanding of intrinsic information in the image, leading to emotion-specific content for $I_s$. After $T-1$ denoising steps, decoder $\mathcal{D}$ produces an emotion-evoked image. We minimize the latent diffusion loss, which is the expected error between the predicted noise by $\epsilon_{\theta}$ and the actual noise $\epsilon$ sampled during training: $L_{noise}=E_{t,y,z,\epsilon(z_t),\epsilon\sim N(0,1)}\left\| \epsilon - \epsilon_{\theta}(z_{t},t,z,\tau_{\theta}(y))\right\| ^{2}$.

\noindent \textbf{Alignment Loss between Model and Human Judgements.} A single emotion term lacks the depth to convey the nuanced causes underlying emotions, posing a challenge for diffusion models to grasp the diagnostic information triggering emotions. In contrast to existing text-to-image models, EmoEditor emulates human reasoning by aligning its thought process with text instructions in our EmoPair dataset, which leads the transition from the source emotion to the target emotion. To achieve this, we introduce an alignment loss in a neuro-symbolic space, between emotion embeddings $e$ from $\tau_{\theta}$ and text embeddings $c$ of instructions in each image pair of our EmoPair dataset with pre-trained CLIP. Specifically, the alignment loss is formulated as the inverse of cosine similarity between $e$ and $c$: $L_{emb}=1-\cos(e,c)$.
Overall, we conduct joint training of $\tau_{\theta}$ and $\epsilon_{\theta}$ with the following loss: $L_{total}=L_{noise}+\lambda L_{emb}$, where $\lambda = 0.5$ represents the weight for balancing different losses. 

Importantly, we differentiate our alignment loss from the losses typically employed in classical language models, which focus on predicting specific word tokens. Our approach loosens the constraint of precisely predicting the exact text instructions. This flexibility empowers our model to explore a wider array of editing solutions, that can augment the source emotions of human viewers to target emotions. Consequently, our model learns to reason in a neuro-symbolic manner akin to humans and preserves its creativity and capacity for exploring novel editing solutions.

\noindent \textbf{Iterative Emotion Inference (IEI).}
Editing emotions is a complex process often requiring iterative image edits. To address this, we introduce a recurrent emotion critic process during inference. This critic iteratively infers the evoked emotions from generated images while assessing their structural coherence and semantic consistency. Specifically, we employ the emotion predictor $\mathcal{P}$ as the critic. 

In the first iteration of image generation, the source image $I_s$ serves as both the input and condition image, producing latent variables $z$ and $z_t$. These, along with the emotion vector $e$, are fed to the EmoEditor to obtain the generated image $\hat{I}$. The critic evaluates the generated image quality based on two criteria: (1) the structural similarity (SSIM) between $\hat{I}$ and $I_s$ falls within 0.5-0.8, and (2) the predicted emotion by $\mathcal{P}$ on $\hat{I}$ matches the target emotion, with a confidence level exceeding 0.6. If both criteria are met, EmoEditor ceases image generation; otherwise, it continues to generate new images.
In subsequent iterations, we use $\hat{I}$ as the input image, generating its noisy latent variable $\hat{z}_t$, with $I_s$ remaining as the condition. We use $\mathcal{P}$ to predict the emotional state of $\hat{I}$ as the new source emotional state and calculate the new emotion editing direction to update the emotion embedding $e$. Based on these new inputs, EmoEditor generates a new edited image for the critic. This process iterates until both criteria are satisfied, or the number of critic iterations exceeds a predefined limit of 30. In the latter case, we select the generated result with the highest confidence level predicted by $\mathcal{P}$ given the target emotion among all past iterations.
See \textbf{Sec S2.5} for more implementation details.

\vspace{-2mm}
\section{Experiment}
\vspace{-2mm}
\label{sec:Experiment}

Ideally, with source images and their classified emotions, our goal is to generate images evoking the other 7 target emotions. Considering emotional valence, we propose two tasks: within-valence emotion editing and cross-valence emotion editing. This division serves two purposes. Firstly, cross-valence emotion editing is more common in real-world scenarios, like enhancing positive emotions in stressful environments or creating fear in lively settings during Halloween. Secondly, images of the same valence may contain subtle differences challenging for humans to discern. For instance, an amusement park scene might evoke a mixture of excitement and amusement. 
We randomly select test images from EmoSet \cite{yang2023emoset}, ensuring they differ from those in our EmoPair dataset and that each source emotion category is evenly distributed. We then evaluate the competitive baselines and our EmoEditor on the cross-valence emotion editing task, with a total of 2,016 test images. We also present the results of applying our EmoEditor to more challenging emotion editing scenarios. To ensure a fair comparison, the input image size is standardized to 224$\times$224 across all methods.

\noindent \textbf{Baselines.}
We compare our EmoEditor with nine methods: 
(1) Color-Transfer (CT) \cite{pitie2007automated}; 
(2) Neural-Style-Transfer (NST) \cite{gatys2015neural}. 
Both methods require an additional reference image from the target emotion category as input. We randomly select these reference images from the EmoSet dataset. We also include representative text-driven image editing models 
(3) Csty \cite{kwon2022clipstyler}; 
(4) CtrlNet \cite{zhang2023ControlNet}
(5) SDEdit \cite{meng2021sdedit}; 
(6) Ip2p \cite{brooks2023instructpix2pix} and 
(7) AIF \cite{weng2023affective}. 
We provide these models with the target emotion category as a one-word text prompt, such as ``excitement". 
We also compare with 
(8) EmoEdit \cite{yang2025emoedit}, which performs emotion-driven image editing by retrieving semantically similar examples from an emotion dictionary. 
In addition, we concatenate existing large language and vision models and introduce the baseline 
(9) ``Large Model Series" (LMS). This includes GPT-4o \cite{hurst2024gpt} for image captioning, followed by GPT-o4 \cite{openai2025o4} for generating reasoning-based text instructions, and Ip2p for image editing based on the instructions (See \textbf{Sec S2.2}).

\noindent \textbf{Human Psychophysics Experiment.}
We evaluate the generated results of all methods on Amazon Mechanical Turk (MTurk) \cite{turk2012amazon}. 
We recruit 136 participants, with each participant undergoing 180 trials, yielding a total of 24,480 trials. 
All the experiments are conducted with the subjects' informed consent and according to protocols approved by the Institutional Review Board of our institution. All participants are properly compensated. In each trial, participants engage in a ``two-alternative forced choice" task. They are presented with two image stimuli and must select the one that most strongly evokes the target emotion in them. The two image stimuli consist of a result generated by our EmoEditor and a randomly sampled result from the baselines introduced above. 
The trial presentation order and the binary choices for each trial are randomized. To assess the quality of data collection, we introduce control trials and use them as filtering criteria. We discard the data from those people who fail the control trials at least 17\% of the time. See \textbf{Sec S2.3} for more details.

\noindent \textbf{Evaluation Metrics.} 
While CLIP \cite{radford2021CLIP} assesses text-image consistency, it does not fully capture emotional complexity \cite{widhoelzl2024decoding}. Thus, we introduce \textbf{Emotional Structure Matching Index (ESMI)}: $ESMI = \alpha \times S_{str} + (1 - \alpha) \times S_{emo}$ where $\alpha$ is a weighting coefficient and set to 0.5. ESMI evaluates emotion-evoked image generation from two aspects: evoking specific emotions and maintaining structural and semantic consistency between the source and generated images. 
\textbf{Evoking Targeted Emotions.}
To assess how well the generated image expresses the target emotion, we use $\mathcal{P}$ to extract emotional distributions $e_s$ and $e_g$ from the source and generated images respectively. Next, we calculate the Kullback-Leibler Divergence between these distributions and the target emotion’s one-hot encoding $e_{oh}$, we get $D_{src} = KLD(e_s, e_{oh})$ and $D_{gen} = KLD(e_g, e_{oh})$. Ideally, the generated image should emotionally be closer to the target, that is, $D_{gen} < D_{src}$. To avoid the influence of outliers or extreme values, we define the raw emotional improvement as $\Delta D = \max(0, D_{src} - D_{gen})$. To further normalize this improvement relative to the original emotional distance, we compute the emotional evocation score: $S_{emo} = \Delta D / (D_{src} + \epsilon)$ where $\epsilon$ is a small constant to prevent division by zero. This score ranges from 0 to 1, with higher values indicating a greater emotional shift toward the target emotion relative to the source. 
\textbf{Preserving Original Structures.}
We define two regions: the Emotion-Evoking Region ($R_{emo}$), which is responsible for triggering the source emotion (e.g. fire for anger) and is expected to be altered, and the Emotion-Neutral Region ($R_{neu}$), which should remain unchanged. We measure structural consistency by computing the pixel-level L1 differences between the source and generated images in these regions, denoted as $L_{emo}$ and $L_{neu}$. 
The structural preservation score is calculated as: $S_{str} = L_{emo} / (L_{emo} + L_{neu} + \epsilon)$ where $\epsilon$ is a small constant to avoid division by zero. This score reflects the proportion of changes in $R_{emo}$ relative to all changed pixels, with higher values indicating better preservation of neutral areas. 
We adopt two strategies to define $R_{emo}$: 
(1) CAM-based: $R_{emo}$ is obtained by applying Grad-CAM \cite{selvaraju2017gradcam} to $\mathcal{P}$ and binarizing the resulting heatmap. (2) HA-based: $R_{emo}$ is derived by averaging manually annotated emotional regions of human annotators. 
See \textbf{Sec S2.4} for more details.

\begin{table*}[t]
  \centering
  \resizebox{0.8\textwidth}{!}{
  \setlength{\tabcolsep}{7pt} 
  \renewcommand{\arraystretch}{1.3} 
  \begin{tabular}{l|ccccccccccc}
    \hline
    ESMI(\%) \textuparrow  & CT    & NST   & Csty  & CtrlNet  & SDEdit & Ip2p  & AIF  & EmoEdit  & LMS   & Ours \\ 
    \hline
    CAM-based & 28.33 & 47.48 & 34.33 & 32.94 & 35.91 & 27.37 & 29.71 & 32.94  & 35.40 & \textbf{51.56} \\ 
    HA-based & 26.68 & 46.00 & 32.77 & 31.53 & 34.27 & 25.72 & 28.31 & 37.53 & 33.76 & \textbf{49.99} \\
    \hline
  \end{tabular}
  }
  \vspace{-2mm}
  \caption{
      \textbf{Quantitative Evaluation of Generated Images for All Competitive Methods.} 
      The \textbf{CAM-based metric} computes ESMI using $R_{\text{emo}}$, derived by binarizing the Grad-CAM output from the Emo Predictor.  
The \textbf{HA-based metric} computes ESMI using $R_{\text{emo}}$, defined as the average of human-annotated regions.  
      Best is in bold. Larger (\textuparrow) is better.  }
  \vspace{-4mm}
  \label{tab:metric}
\end{table*}

\begin{figure}[t]
    \centering
    \includegraphics[width=0.96\linewidth]{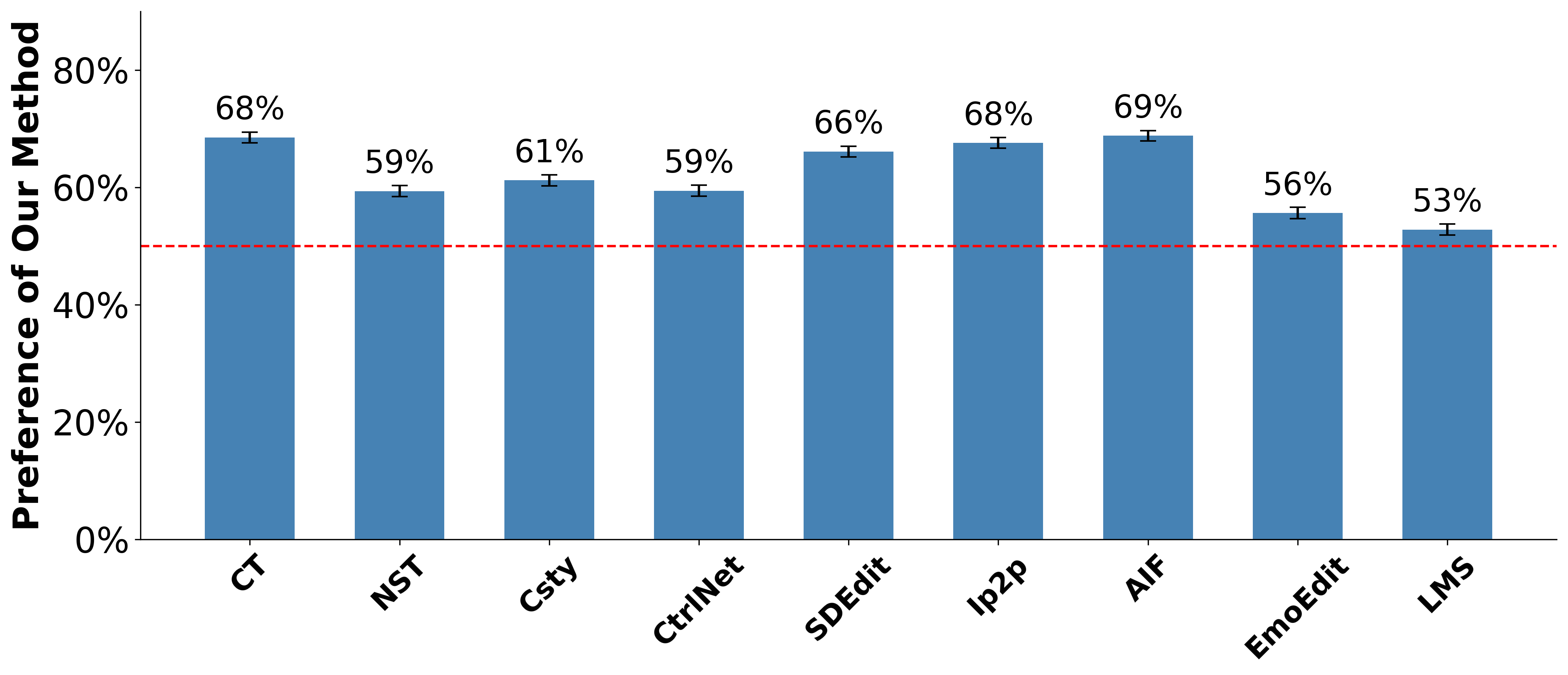}
    \vspace{-2.5mm}
    \caption{
    \textbf{Results from Human Psychophysics Experiments.} The proportions of images that human participants prefer our EmoEditor over all other methods on the x-axis are presented.
        Chance is 50\% (red dotted line). Error bars are standard errors.
    }
    \label{subfig:human_experiment}
    \vspace{-5mm}
\end{figure}

\begin{table}[t]
  \centering
  \resizebox{0.89\linewidth}{!}{
    \begin{tabular}{c|ccc|c|c|c}
        \hline
        \multirow{2}{*}{Variant} & \multicolumn{3}{c|}{Input of $\tau_{\theta}$} & \multirow{2}{*}{IEI} & \multirow{2}{*}{$L_{emb}$} & \multirow{2}{*}{ESMI(\%)\textuparrow} \\ 
        \cline{2-4}
         & text & $e_{oh}$ & $e_{dir}$ & & & \\
        \hline
       1 & \Checkmark & &  &  &    & 5.85  \\
       2 & \Checkmark & & & \Checkmark &    & 50.24 \\
       3 & & \Checkmark & & & \Checkmark     & 7.36  \\
       4 & & & \Checkmark & & \Checkmark     & 8.51  \\
       5 & & \Checkmark & & \Checkmark &    & 47.36 \\
       6 & & \Checkmark & & \Checkmark & \Checkmark &   49.80 \\
      7 (ours) &  & & \Checkmark & \Checkmark & \Checkmark &   \textbf{51.56} \\ 
        \hline
    \end{tabular}
  }
  \vspace{-2mm}
  \caption{
        \textbf{Quantitative Evaluation in CAM-based ESMI for Model Variants of our EmoEditor.} A tick indicates the inclusion of a component in the corresponding model variant, while a blank cell indicates its absence. The last row represents our full EmoEditor model. Best results are in bold. Larger (\textuparrow) is better. 
  }
  \vspace{-4mm}
  \label{tab:abstudy}
\end{table}

\vspace{-2mm}
\section{Results}
\subsection{Evaluation in Cross-Valence Scenarios}
\label{sec:Res_Quantitative}
\textbf{Quantitative Evaluation with SOTA.} 
Figure \ref{subfig:human_experiment} shows human psychophysics experimental results (see \textbf{Sec S3.1} for more results). If all methods generate images that elicit target emotions equally effectively, human participants would make random choices, with a 50\% chance in the ``two-alternative forced choice" tasks. 
Observing Figure \ref{subfig:human_experiment}, we see that human participants consistently prefer the generated results of our EmoEditor over all competitive baselines, indicating its proficiency in evoking target emotions. Complimentary to psychophysics experiments, Table \ref{tab:metric} presents quantitative results in CAM-based and HA-based ESMI, with our EmoEditor achieving the highest scores in both metrics. 
Notably, for each method, the scores remain consistent across the two evaluation metrics, indicating that the emotion predictor $\mathcal{P}$ identifies emotion-evoking regions in a manner aligned with human annotations. This consistency further supports the reliability and validity of the CAM-based ESMI metric.

Our EmoEditor significantly outperforms CT, demonstrating that emotion-evoked image generation requires more than global color adjustments. While NST and Csty can induce negative emotions through distortions and irregular textures, they often reduce image interpretability (see \textbf{Fig S10}). 
CtrlNet and SDEdit produce abstract textures that may evoke emotion but often fail to preserve the source image’s semantics. Human evaluations also prefer EmoEditor over Ip2p, underscoring the limitations of current diffusion models trained for general image generation when applied to emotion-driven tasks. 
Although both AIF and EmoEdit are designed for emotional editing, our EmoEditor achieves higher human preference and ESMI scores. 
AIF applies global filters, resulting in less targeted changes. 
EmoEdit maps image-emotion pairs into an embedding, but the limited and biased emotion dictionary leads to repetitive edits and poor performance on strongly emotional inputs. 
Finally, our EmoEditor outperforms LMS in both human preference and ESMI, suggesting that simply chaining large models may amplify biases and errors. Emotion-evoked generation demands a nuanced understanding of both the source image and target emotion, making end-to-end learning crucial.

\begin{figure}[t]
    \centering
    \includegraphics[width=0.96\linewidth]{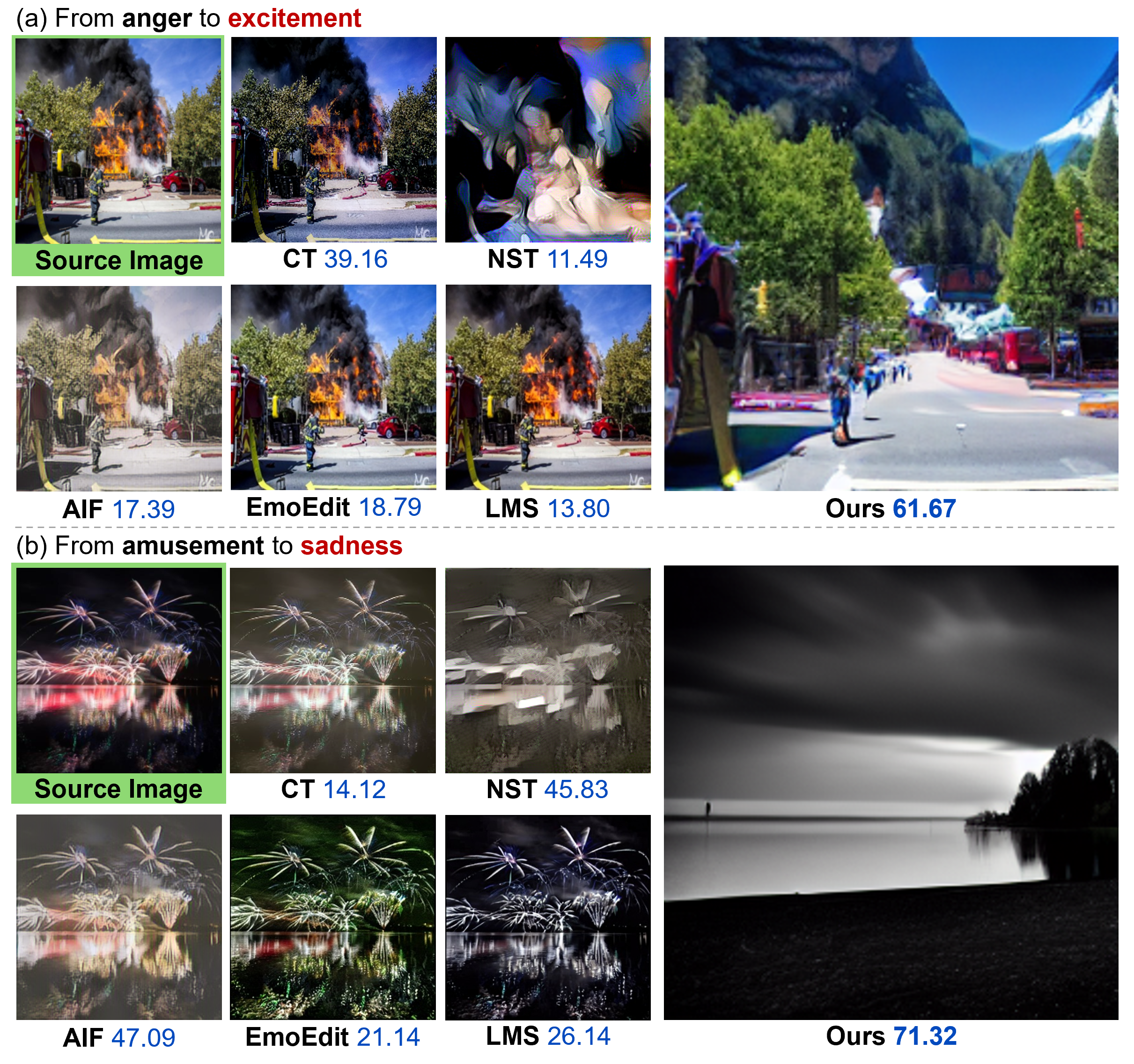}
    \vspace{-2.5mm}
    \caption{\textbf{Visualization of Generated Images from Different Methods.} Target emotion is highlighted in red and source image is framed in green. The blue number below each image is its CAM-based ESMI score, with higher values being better. More examples in \textbf{Sec S3.2}.
 }
    \label{fig:fig9_visualization}
    \vspace{-5mm}
\end{figure}

\begin{figure*}[ht]
    \centering
    \includegraphics[width=0.88\linewidth]{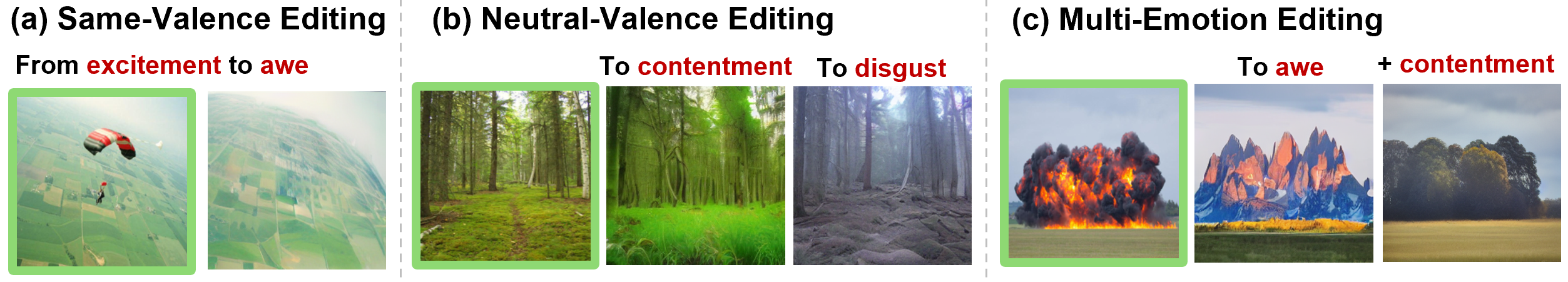}
    \vspace{-2.5mm}
    \caption{\textbf{Our EmoEditor can generalize to more challenging emotion editing scenarios.} (a) \textbf{Same-Valence Editing} highlights our EmoEditor's ability to produce images evoking emotions of the same positive valence as the source images. (b) \textbf{Neutral-Valence Editing} demonstrates how our EmoEditor can transform neutral real-world images to evoke positive or negative emotions. (c) \textbf{Multi-Emotion Editing} demonstrates its capability to generate images that evoke a wider range of emotions, providing more nuanced and diverse visual outputs. 
    Source images are framed in green. Target emotions are in red. 
    }
    \label{fig:fig12_pos2pos}
    \vspace{-4mm} 
\end{figure*}

\noindent \textbf{Model Analysis.} 
We conduct the model analysis and present the results in Table \ref{tab:abstudy}.
In Row 1, the model variant takes only text as input and follows the structure of Ip2p. Although it is fine-tuned on our EmoPair dataset with the pre-trained CLIP model as $\tau_{\theta}$, the performance is the lowest among all variants, indicating that text guidance is insufficient. Adding IEI (Row 2) results in notable improvements over the version without IEI (Row 1), confirming its contribution to enhancing emotional consistency. Next, replacing the text input with a one-hot encoding of the target emotion $e_{oh}$ (Row 3) leads to a better result, highlighting the advantage of structured emotion encoding over text. 
Further improvement is observed when replacing $e_{oh}$ with an emotion editing direction $e_{dir}$ (Row 4), suggesting that incorporating information about the source emotion enables more precise and effective editing. 
Adding the alignment loss $L_{emb}$ (Row 6) improves performance over the variant without it (Row 5), confirming its complementary role in promoting emotionally consistent generation. 
Our final model (Row 7) integrates $e_{dir}$, IEI, and $L_{emb}$, enabling emotion-aware editing based on the source-target gap. This configuration achieves the highest score, demonstrating its superior emotional precision and generative diversity.

\noindent \textbf{Visualization of Emotional Image Generation across Valence.} 
Figure \ref{fig:fig1_teaser} and \ref{fig:fig9_visualization} present visualizations of generated images in cross-valence scenarios. 
CT primarily replicates tonal characteristics from a reference image, often failing to significantly enhance emotions. NST relies on randomly selected references, producing artistic textures aligned with target emotions but losing essential semantic content. AIF, dependent on emotional text descriptions, applies a global filter effect but struggles with localized editing. 
EmoEdit relies on fixed learned queries, leading to limited, context-incongruent edits (e.g., always adding balloons for amusement; Figure~\ref{fig:fig1_teaser}) and failing to edit local emotional cues such as fire flames or fireworks (Figure~\ref{fig:fig9_visualization}). 
Even with a richer vocabulary, such as LMS, which integrates three large models, it still struggles to generate desirable outputs. 
In contrast, our EmoEditor takes only a source image and target emotion as input, yet generates creative and emotionally aligned results while preserving structural and semantic coherence. By learning an emotion editing direction, it understands the source-target gap and applies precise edits. 
In Figure~\ref{fig:fig9_visualization}, where fire and fireworks dominate the original emotion, EmoEditor adapts its edits to the target: extinguishing flames and adding mountains to evoke excitement (a), or removing fireworks and darkening tones to convey sadness (b). In Figure~\ref{fig:fig1_teaser}, it replaces fire flames with  lamps to elicit amusement.
Notably, even when handling similar anger-inducing elements like fire, our EmoEditor produces diverse, context-aware edits tailored to the source image without any manual design, demonstrating a nuanced understanding of emotional cues. 
We also show CAM-based ESMI scores (in blue) for each result. Our EmoEditor achieves the highest scores, demonstrating its ability to accurately identify and effectively modify emotion-evoking regions in source images, striking a balance between evoking the target emotion and preserving the original structure.

\subsection{Generalization to Real-world Scenarios}
\label{sec:Generalization}

Editing images with the same emotional valence is more challenging due to subtle emotional cues. Figure \ref{fig:fig12_pos2pos}a demonstrates EmoEditor's ability to handle such tasks, as it removes a parachutist to reveal a breathtaking landscape, shifting excitement to awe. 
In real-world scenarios, some naturalistic images may evoke mixed emotions or no strong emotional responses at all. We challenge our EmoEditor to generate emotion-evoked images using neutral-valence source images randomly selected from the MSCOCO dataset \cite{lin2014microsoft}. These images depict scenes of everyday life with neutral emotions. EmoEditor applies positive and negative emotion transformations to these source images. It enhances the lushness of a forest to evoke contentment and transforms it into a barren landscape for disgust (Figure \ref{fig:fig12_pos2pos}b). Notably, EmoEditor generates these results purely from its creativity, \textit{without} any human-annotated editing instructions or scene interpretations.

Furthermore, EmoEditor can manage multi-emotion editing with the use of emotion encoding as input. We extend the single-target emotion scenario by randomly adding one additional emotion of the same valence as the target and then use EmoEditor for image generation. In Figure \ref{fig:fig12_pos2pos}c, EmoEditor generates a majestic mountain in the flame area of the source image to evoke awe. After adding contentment, it enriches the area with dense forests.

\section{Discussion}
\label{sec:discussion}
We introduce the challenging problem of emotion-evoked image generation and propose EmoEditor, an image diffusion model that understands emotional cues, creates implicit editing instructions aligned with human decisions, and manipulates image regions to evoke emotions, while maintaining coherent scene structures and semantics. Moreover, we contribute EmoPair dataset for model training. To benchmark all methods, we introduce a new evaluation metric that effectively measures the balance between evoking target emotions and maintaining the structural and semantic integrity of the source image. 

While EmoEditor excels in quantitative performance and visual results, it struggles with fine details on faces (See \textbf{Sec S3.4}). While we adopt the 8-category Mikels model, real-world emotions are far more nuanced and complex. In particular, differentiating between positive emotions tends to be more challenging than distinguishing negative ones. Additionally, EmoEditor carries risks of misuse, such as evoking undesirable emotions, which could harm mental well-being or mislead public sentiment. All of these aspects warrant further investigation in future research.

\appendix
\onecolumn

\setcounter{table}{0}   
\setcounter{figure}{0}
\setcounter{section}{0}
\setcounter{equation}{0}
\renewcommand{\thetable}{S\arabic{table}}
\renewcommand{\thefigure}{S\arabic{figure}}
\renewcommand{\thesection}{S\arabic{section}}
\renewcommand{\theequation}{S\arabic{equation}}


\section{EmoPair Dataset}
\label{sec:SP_Emopair}

In response to the absence of an image dataset specifically designed for emotion-evoke image generation, we introduce EmoPair, comprising two distinct subsets: the EmoPair-Annotated Subset (EPAS), encompassing 331,595 image pairs sourced from Ip2p \cite{brooks2023instructpix2pix} and annotated with emotion labels; and the EmoPair-Generated Subset (EPGS), featuring 6,949 pairs generated through text instructions specifying target emotions.

For the EmoPair-Annotated Subset (EPAS), we use $\mathcal{P}$ to categorize the source and target images from the Ip2p dataset into eight different emotions. We remove samples where the emotional labels of the source and target images are consistent, then use the emotion labels to augment the Ip2p dataset, generating 331,595 image pairs. 

\begin{figure}[ht]
    \centering
    \includegraphics[width=0.6\linewidth]{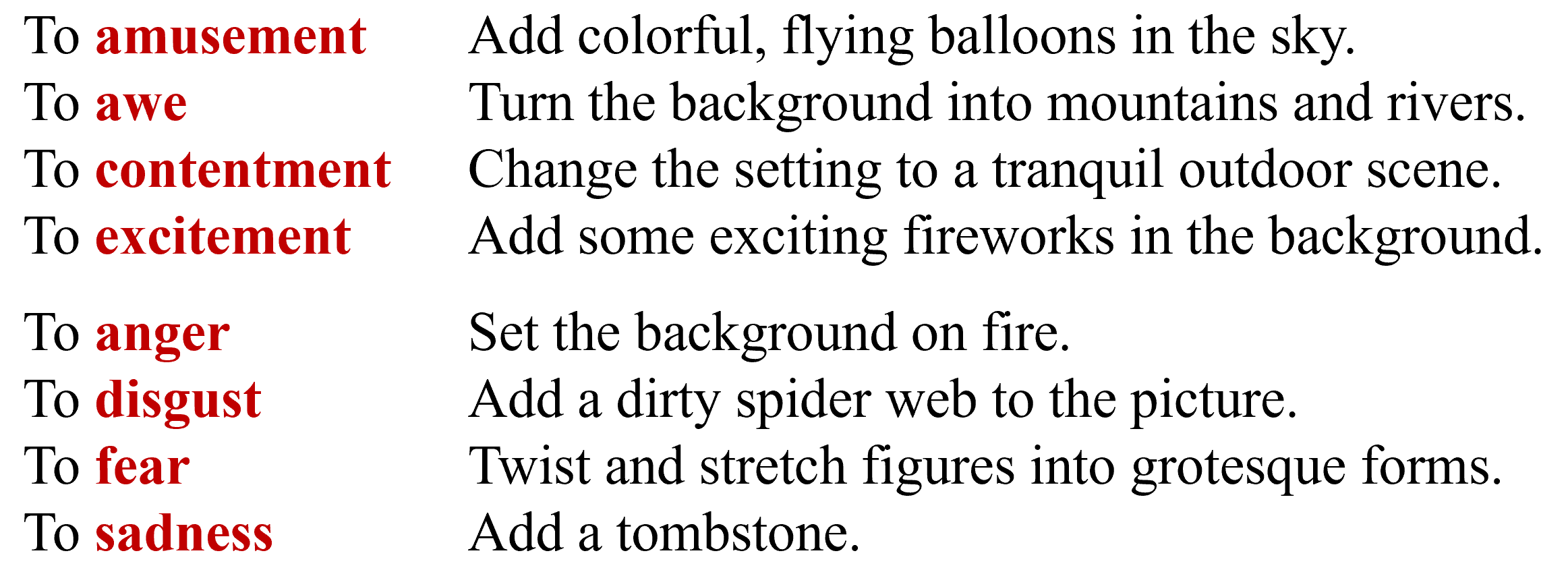}
    \caption{\textbf{Instruction examples of EmoPair-Generated Subset (EPGS).} For each emotion category, we retained the top ten text instructions based on the rankings determined by human annotators according to the efficacy of each emotion category. We utilized these instructions for Ip2p in image editing to generate target images capable of evoking the desired emotions, thereby constructing our EPGS.
    }
    \label{fig:SP_10instructions}
\end{figure}

For the EmoPair-Generated Subset (EPGS), we formulated 50 general instructions that are agnostic to source images, prompting transitions to desired emotions across 8 categories using GPT-3 \cite{brown2020GPT3}. Human annotators then ranked these instructions based on efficacy within each emotion category, ultimately retaining the top ten. Figure \ref{fig:SP_10instructions} illustrates examples of these instructions. 

We employ the following selection criteria to control the quality of generated image pairs of EPGS: (1) Using our emotion predictor $\mathcal{P}$, we analyze the generated images and only select those with a Top-1 classification confidence over 90\% for the target emotions. (2) To ensure the preservation of similar scene structures, we utilize  Structural Similarity Index (SSIM) \cite{wang2004SSIM} and Learned Perceptual Image Patch Similarity (LPIPS) \cite{zhang2018LPIPS} for filtering the remaining outcomes. SSIM measures structural similarity between images, while LPIPS quantifies perceptual differences. Specifically, we require the generated images $\hat{x}$ and the source image $x$ to meet the conditions below: $0.3 < SSIM(x, \hat{x}) < 0.6$ and $LPIPS(x, \hat{x}) > 0.1$. Ultimately, EPGS retains 6,949 image pairs.

To ensure the quality of EPGS, we have human annotators re-annotate 300 images. The annotators view both the source and edited images, and we ask them to select the one that better evokes the target emotion. Since we need the target image to evoke the desired emotion more effectively than the source image, if the annotators choose the source image, we swap the source and target images in the original dataset to create a new subset. Each image is reviewed by three annotators, and the final subset is determined by majority vote. We then use this subset to fine-tune our EmoEditor.

\begin{figure*}[ht]
    \centering
    \begin{minipage}{0.49\textwidth}
        \centering
        \includegraphics[width=\linewidth]{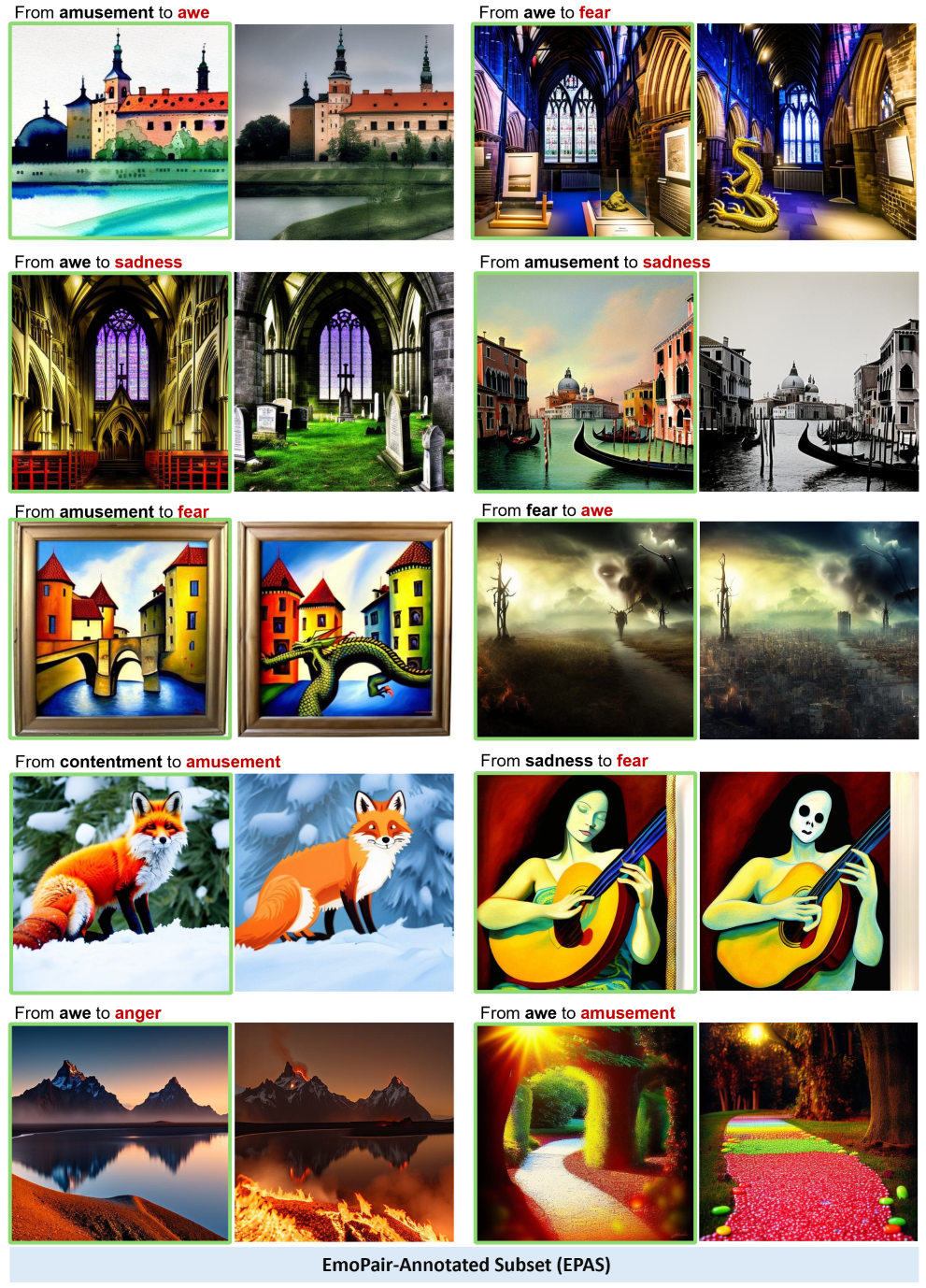}
        \caption{\textbf{Sample of EmoPair-Annotated Subset (EPAS).} On the left side of each pair of images is the source image (framed in green), while the right side shows the target image. The emotion labels for the source and target images (highlighted in red) are indicated above the images.}
        \label{fig:SP_EPAS}
    \end{minipage}\hfill
    \begin{minipage}{0.49\textwidth}
        \centering
        \includegraphics[width=\linewidth]{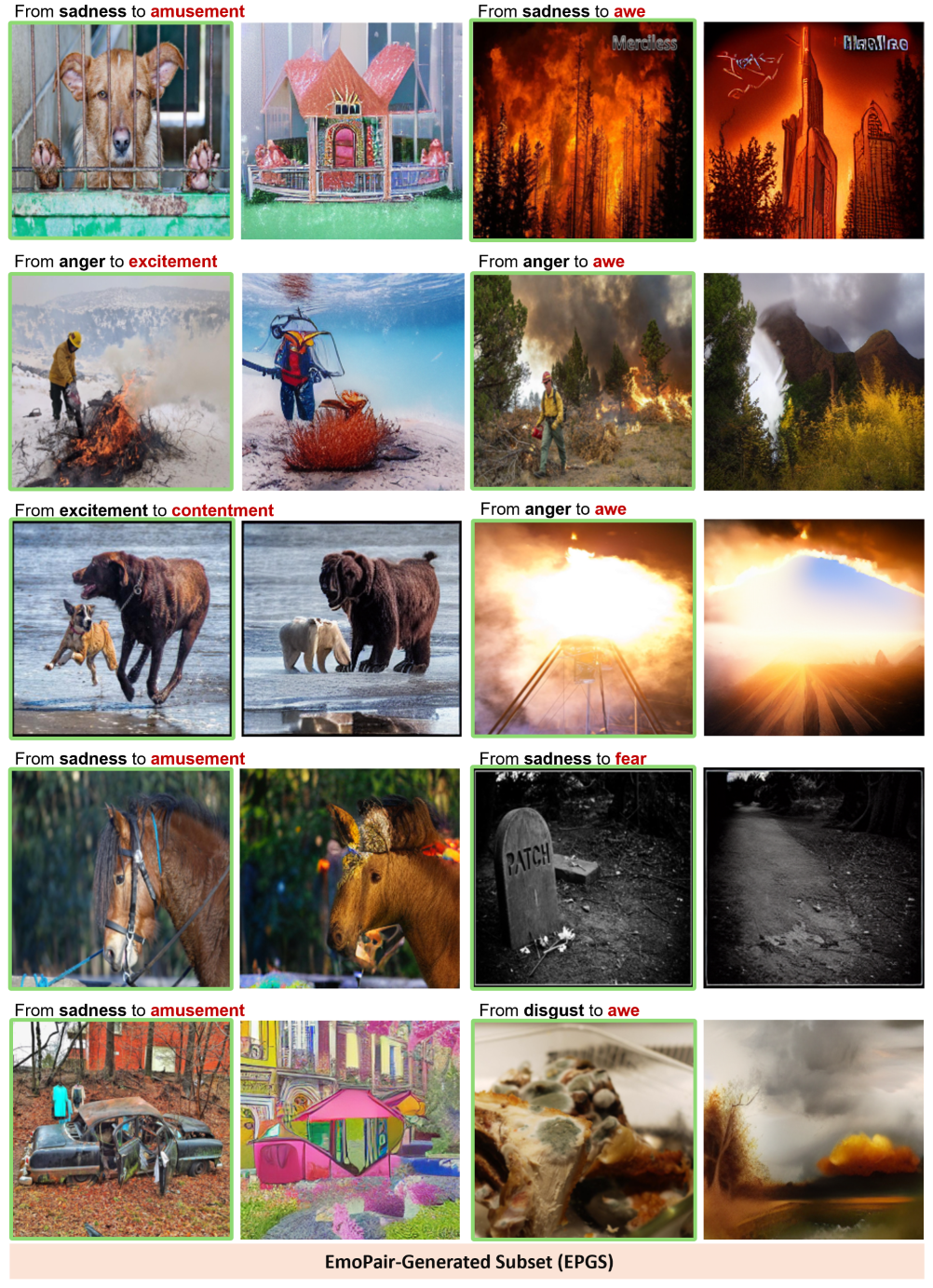}
        \caption{\textbf{Sample of EmoPair-Generated Subset (EPGS).} On the left side of each pair of images is the source image (framed in green), while the right side shows the target image. The emotion labels for the source and target images (highlighted in red) are indicated above the images.}
        \label{fig:SP_EPGS}
    \end{minipage}
\end{figure*}

To provide a comprehensive overview of our dataset, additional image pair examples are presented in Figure \ref{fig:SP_EPAS} and \ref{fig:SP_EPGS}.

\section{Experiment}
\label{sec:SP_exp}
\subsection{Emotion Encoder}
\label{sec:SP_emoenc}
\begin{figure}[ht]
    \centering
    \includegraphics[width=0.7\linewidth]{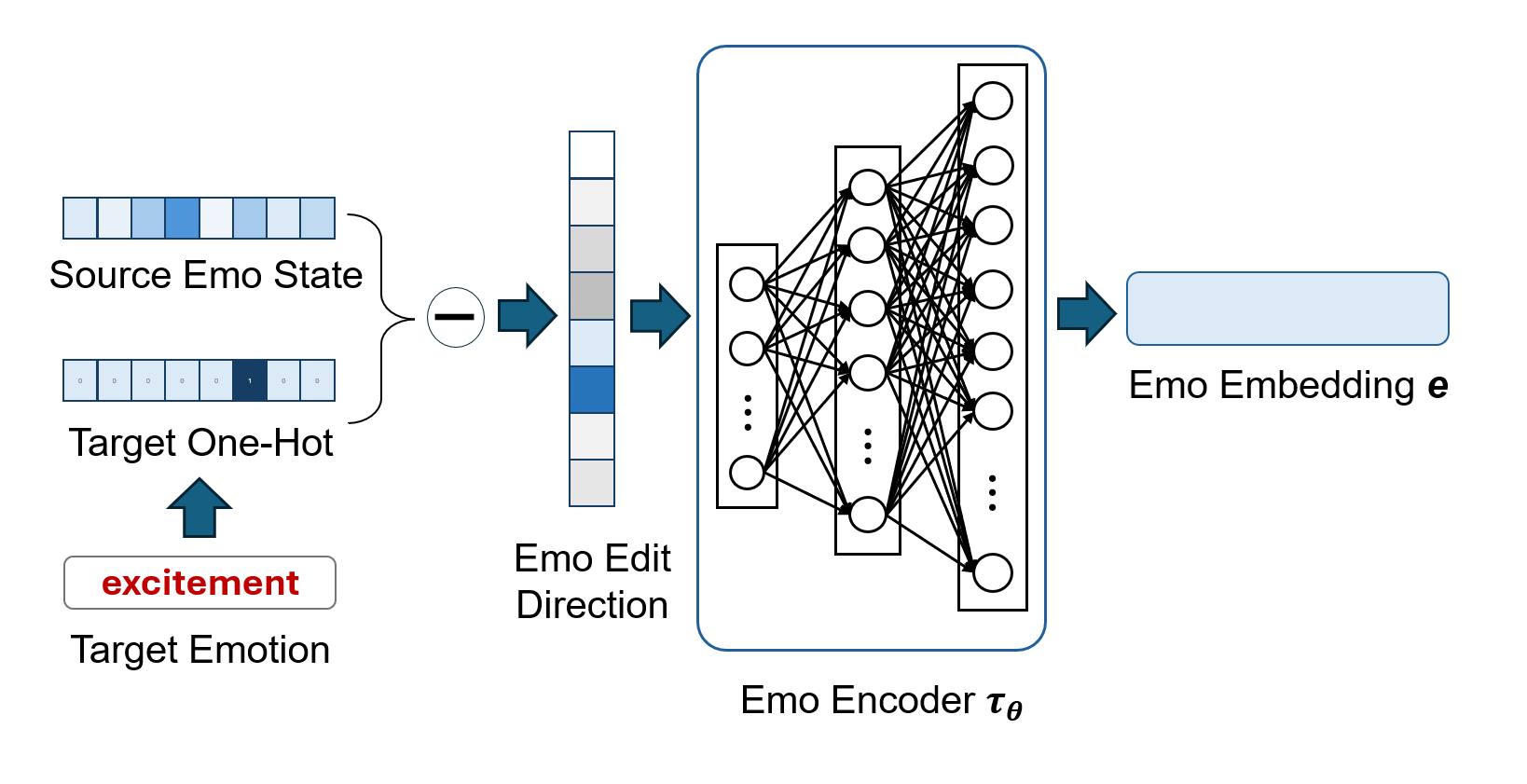}
    \caption{\textbf{The network structure of our emotion encoder $\tau_{\theta}$.}
    }
    \label{fig:SP_emo_enc}
    \vspace{-6mm}
\end{figure}

Figure \ref{fig:SP_emo_enc} shows the network structure of our emotion encoder $\tau_{\theta}$. We represent the target emotion as a one-hot encoding $e_{oh}$ and use the emotion predictor $\mathcal{P}$ to calculate the emotional state of the source image $e_s$. Then, we compute the emotion editing direction $e_{dir}$ by calculating the difference between $e_{oh}$ and $e_s$. Our Emotion Encoder $\tau_{\theta}$ is a fully connected network designed for transforming emotion editing direction $e_{dir}$ into the structured emotion embedding $e$. The architecture of the model is composed of a series of fully connected layers that progressively increase the dimensionality of the input. 

The network begins with an input layer of size 8, which is then passed through a sequence of fully connected layers with increasing sizes: 256, 512, and 768 neurons, respectively. Each of these layers is followed by a ReLU activation function to introduce non-linearity and improve the model's learning capacity. The final linear layer projects the output to a dimension of $77\times768$, effectively structuring the embedding to fit the input size of our denoising network $\epsilon_{\theta}$.

\vspace{-4mm}

\subsection{Baseline: Large Model Series}
\label{sec:SP_LMS}
Emotion-evoked image generation involves three key steps: image understanding, instruction generation, and instruction-based image editing. To tackle these, we concatenate existing large language and vision models and introduce the baseline ``Large Model Series" (LMS). This includes GPT-4o \cite{hurst2024gpt} for image captioning, followed by GPT-o4 \cite{openai2025o4} for generating reasoning-based text instructions, and Ip2p for image editing based on the instructions.

\begin{figure}[ht]
    \centering
    \includegraphics[width=0.6\linewidth]{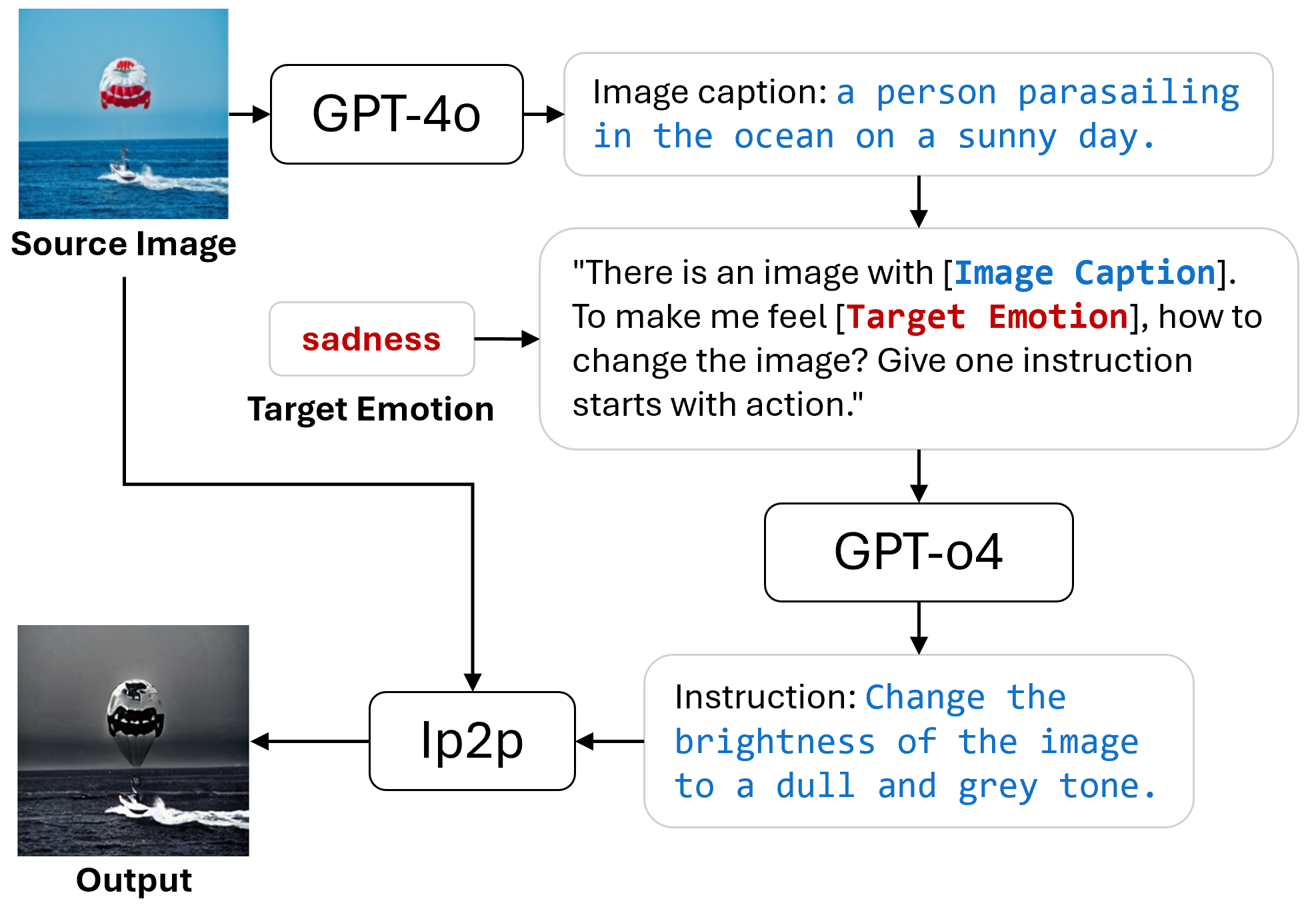}
    \caption{\textbf{Workflow of Large Model Series (LMS).} We decompose emotion-evoked image generation into three steps: employing GPT-4o for image understanding, utilizing GPT-o4 for instruction generation, and employing Ip2p for image editing. We refer to this approach of chaining multiple large models as the Large Model Series.
    }
    \label{fig:SP_LMS}
\end{figure}

Figure \ref{fig:SP_LMS} shows the workflow of LMS. First, we employ GPT-4o to comprehend the source image, generating an image caption corresponding to the source image. Subsequently, based on the generated caption, we utilize the sentence structure ``There is an image with [Image caption]. To make me feel [Target Emotion], how to change the image? Give one instruction that starts with an action." to query GPT-o4. Following this, GPT-o4 generates an instruction to guide Ip2p in editing the source image, resulting in the final output.

\subsection{Human Psychophysics Experiment}
\label{sec:SP_human}

\begin{figure*}[ht]
	\centering
	\includegraphics[width=0.8\linewidth]{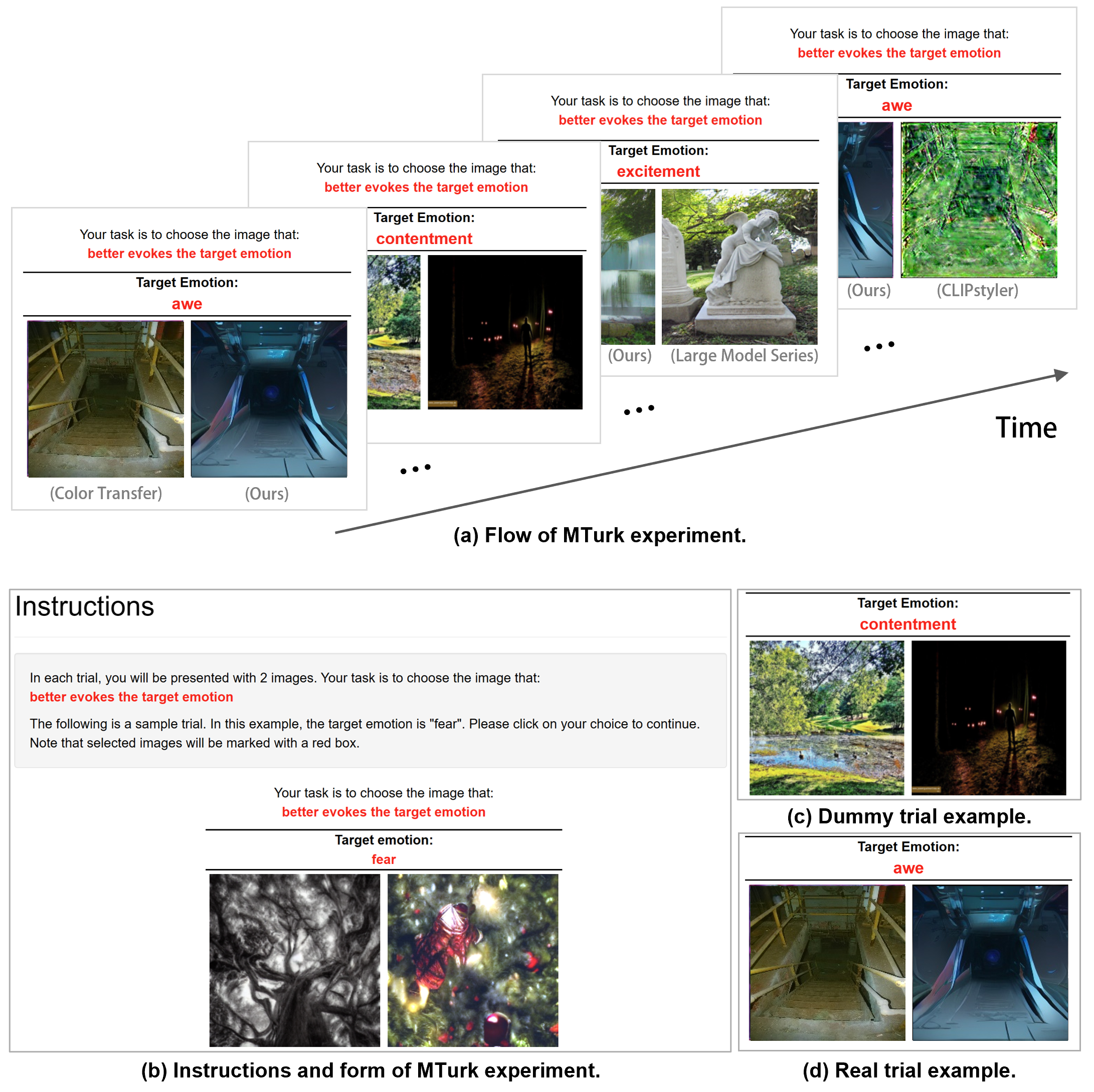}\vspace{-3mm}
	\caption{\textbf{MTurk Experiment.} (a) Schematic of MTurk experiment. Participants are presented with a set of images and a target emotion. They must choose between two images, one generated by our model and the other by 9 state-of-the-art methods, selecting the one that more effectively evokes the target emotion. (b) Instructions of the MTurk experiment. (c) Dummy trial example. We chose six image pairs with prominent emotional distinctions as benchmarks to assess participants' comprehension of the task and their attentiveness to the experiment. (d) Real trial example.
 }
	\label{fig:SP_MTurk}
\end{figure*}

Figure \ref{fig:SP_MTurk}(a) shows the MTurk experiment schematic and Figure \ref{fig:SP_MTurk}(b) shows the experiment instruction pages given to the participants. To control the quality of data collected, we have implemented preventive measures to screen participants:

(a) Each participant undergoes 6 randomly dispersed dummy trials within the real experiments. To assess if participants are making random selections, we use 6 image pairs with prominent emotional differences as references. Considering individual emotional response variability, participants are allowed a maximum of one incorrect choice in these dummy trials. Subjects exceeding an error rate of 1/6 are excluded, resulting in 24,480 trials. The outcomes of the 6 dummy trials are also excluded from the final result analysis. Figure \ref{fig:SP_MTurk}(c) provides an example of a dummy trial.

(b) Each participant can only take part in the experiment once.

(c) Image pairs used in the entire experiment are drawn from the 2,016 result pairs generated by our model and other SOTA methods.

Participants view randomly sampled pairs from these 2,016 pairs, and all trials are presented in a random order. The order of the two images for selection is also randomized. An example trial in the real experiment is illustrated in Figure \ref{fig:SP_MTurk}(d).

\subsection{Evaluation Metrics: Structural Score}
Emotion-evoked image generation needs to be evaluated from both emotional induction and structural preservation perspectives. Traditional metrics, such as SSIM, are inadequate for structural preservation evaluation because they do not consider emotional cues. The regions in the image that evoke emotions are called Emotion-Evoking Regions ($R_{emo}$). These areas need to be altered to change the source emotion and induce the target emotion. Conversely, some regions in the image lack significant emotional cues and are referred to as Emotion-Neutral Regions ($R_{neu}$). These regions should be preserved as much as possible during the editing process.

\begin{figure*}[ht]
    \centering
    \includegraphics[width=0.6\linewidth]{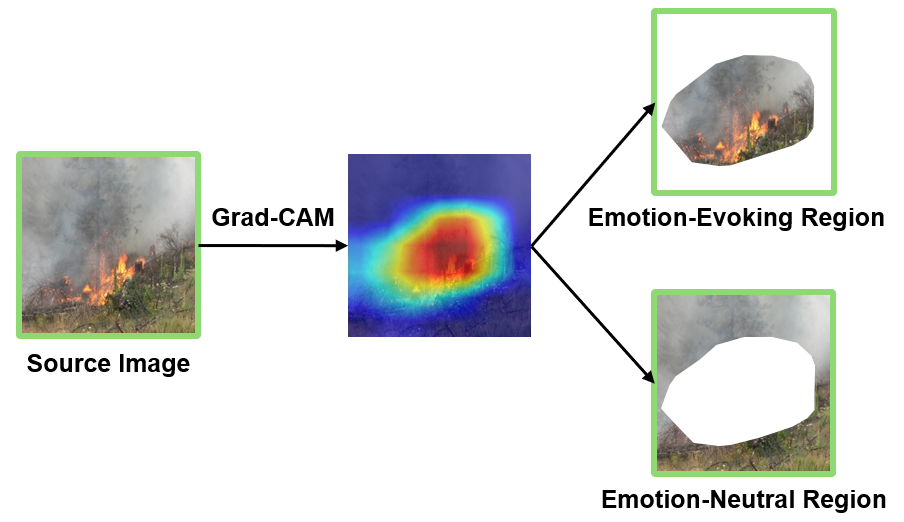}
    \caption{\textbf{Visualization Explanation of the Emotion-Evoking and Emotion-Neutral Regions.} The source image is framed in green. 
    Using Grad-CAM \cite{selvaraju2017gradcam} with $\mathcal{P}$, we binarize Grad-CAM maps at a 0.5 threshold to identify Emotion-Evoking Regions (valued at 1) and Emotion-Neutral Regions (valued at 0) on source images.
    }
    \label{fig:SP_CAM}
\end{figure*}

Grad-CAM \cite{selvaraju2017gradcam} can visualize which parts of the image most influence the model's decision. Therefore, we use Grad-CAM with the emotion predictor $\mathcal{P}$ to identify Emotion-Evoking Regions ($R_{emo}$) in the source images. Figure \ref{fig:SP_CAM} shows the visualization of the Emotion-Evoking and Emotion-Neutral Regions. Using Grad-CAM \cite{selvaraju2017gradcam} with $\mathcal{P}$, we binarize Grad-CAM maps at a 0.5 threshold to identify Emotion-Evoking Regions (valued at 1) and Emotion-Neutral Regions (valued at 0) on source images.

To evaluate changes, we calculate pixel-level L1 differences between the source and generated images for these regions, denoted as $L_{emo}$ and $L_{neu}$. The structural score is: $S_{str} = L_{emo} / (L_{emo} + L_{neu} + \epsilon)$ where $\epsilon$ is a small constant to avoid division by zero. This score reflects the proportion of changes in $R_{emo}$ relative to all changed pixels, with higher values indicating better preservation of neutral areas.

\begin{figure*}[ht]
    \centering
    \includegraphics[width=0.4\linewidth]{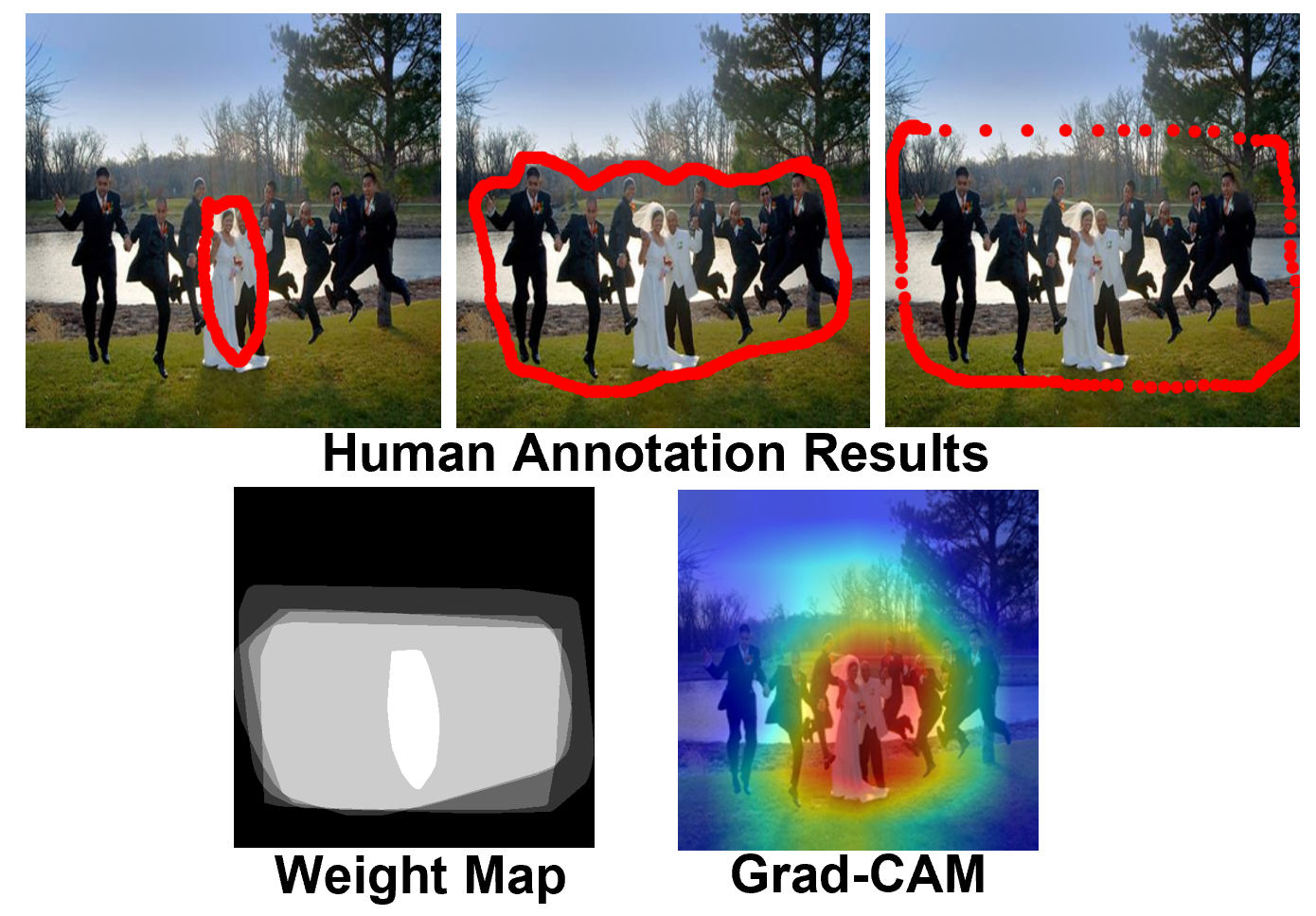}
    \caption{
               \textbf{Emotion-Evoking Region Annotated by Human.} 
        Each image has been annotated by at least three human annotators who identified the emotion-evoking regions. We averaged all the annotations and binarized the result to generate a human-annotated weight map. Row 2 presents both the weight map and the heatmap computed using Grad-CAM, showing that the two are largely consistent.
    }
    \label{fig:SP_anno}
\end{figure*}

Considering that Grad-CAM may be inaccurate or unstable due to the limitations of the emotion predictor $\mathcal{P}$, we invited 46 participants to annotate the emotion-evoking regions of 504 source images. Each worker annotated 50 trials, resulting in a total of 2,300 trials. 
Additionally, we included 6 dummy trials, which were the same as those used in our human psychophysics experiment (Figure \ref{fig:SP_MTurk}(c)), to ensure annotation quality. 
Results from workers who failed more than once on the dummy trials were discarded. Ultimately, each image was annotated by at least three individuals. We then averaged all the annotations and binarized the results to generate a human-annotated weight map. Figure \ref{fig:SP_anno} shows example annotations and a comparison with Grad-CAM.

\subsection{Implementation Details}
\label{sec:SP_ImplementationDetails}
Following \cite{wang2022pretraining}, we initialize the weights of $\mathcal{E}$, $\mathcal{D}$, and $\epsilon_{\theta}$ with the pre-trained Ip2p \cite{brooks2023instructpix2pix} weights. Throughout the training process, we maintain the fixed parameters of $\mathcal{E}$ and $\mathcal{D}$, focusing on training $\tau_{\theta}$ and $\epsilon_{\theta}$. The frozen emotion predictor $\mathcal{P}$ is only used during inference. We conduct experiments with NVIDIA RTX A5000 GPUs, implementing our PyTorch-based framework using the ADAM optimizer. The max time step $T$ is set to 1,000. 


\begin{figure*}[ht]
    \centering
    \includegraphics[width=0.7\linewidth]{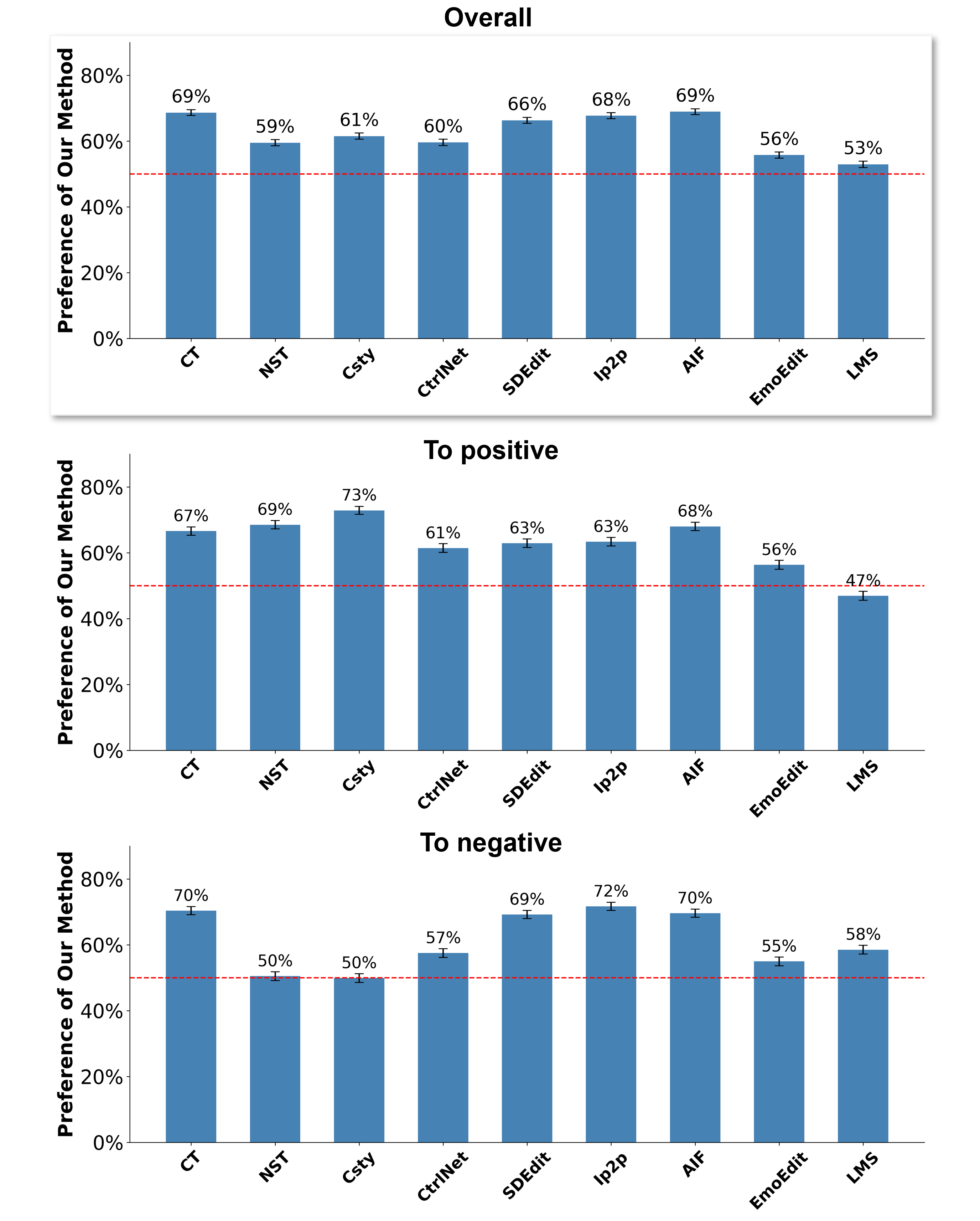}
    \caption{
    \textbf{Preference of Our Method Over SOTA Methods.} 
    The first row shows the overall results, the second row only considers the trials with transitions from negative to positive emotions, and the third row focuses on the trials with transitions from positive to negative emotions. Chance is 50\% in the red dotted line. Error bars are standard errors.
    }
    \label{fig:SP_bar}
\end{figure*}

\section{Results}
\subsection{Quantitative Evaluation in Cross-Valence Scenarios} 
\label{sec:SP_Quantitative}
We assess the generated outputs of all methods using Amazon Mechanical Turk (MTurk)\cite{turk2012amazon} (Online). We recruit 136 participants, with each participant undergoing 180 trials, yielding a total of 24,480 trials. Figure \ref{fig:SP_bar} illustrates the preference of our method over other state-of-the-art methods, demonstrating the superiority of our EmoEditor in producing results that evoke the target emotion compared to other state-of-the-art methods.

From the overall results (Row 1), we see that human participants consistently prefer the generated results of our EmoEditor over all competitive baselines. 
Results from the ``To positive" (Row 2) and ``To negative" (Row 3) indicate that Neural-Style-Transfer (NST) \cite{gatys2015neural} and CLIP-Styler (Csty) \cite{kwon2022clipstyler} perform comparably to our method in generating images that evoke negative emotions, but struggle with positive outcomes. This indicates that negative emotions are easily influenced by factors such as chaotic textures and overall tones, whereas positive emotions are more challenging and require an understanding of the source image along with effective adjustments. Our EmoEditor significantly outperformed Color Transfer (CT) \cite{pitie2007automated} and AIF \cite{weng2023affective} in all scenarios. This confirms that emotion-evoked image generation involves more than just global color and brightness adjustments. 
Additionally, our EmoEditor surpasses SDEdit \cite{meng2021sdedit} and Ip2p \cite{brooks2023instructpix2pix} in generating results evoking positive emotions, while the improvement in generation performance is even more significant in generating results evoking negative emotions. This highlights challenges in current diffusion models for understanding and generating emotion-evoked images and indirectly confirms the greater challenge of generating positive emotion-evoked images compared to negative ones. 

For both positive and negative emotion generation, our method is preferred over EmoEdit in human evaluations. Furthermore, the performance of our EmoEditor and the Large Model Series (LMS) is comparable. While LMS achieves slightly higher scores in positive emotion generation, the margin is narrow and largely influenced by its reliance on powerful external reasoning modules. In contrast, our method offers a compact and fully end-to-end solution that generalizes well across both positive and negative emotions, without the need for multi-stage processing or large-scale language models. Moreover, these language models are often proprietary and equipped with safety constraints, which limit their applicability in privacy-sensitive contexts.

\begin{figure*}[ht]
    \centering
    \includegraphics[width=0.99\linewidth]{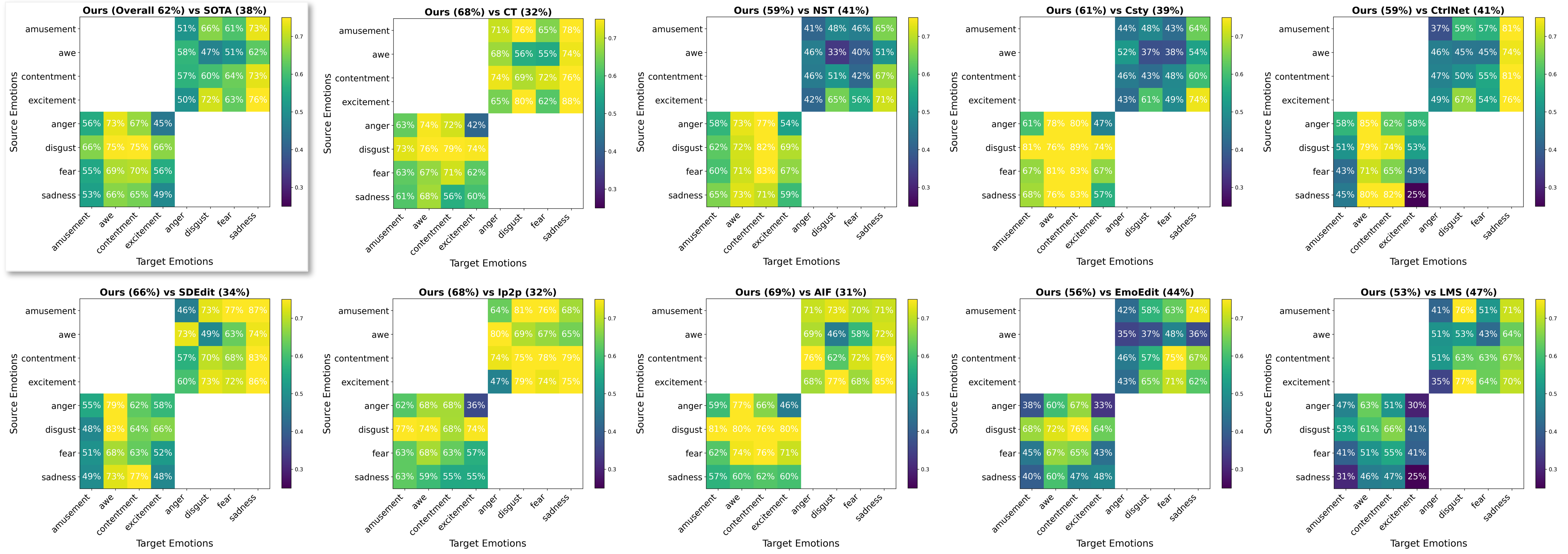}
    \caption{
    \textbf{Confusion matrix of different methods.} The confusion matrix reflects the human preference for our EmoEditor over other methods based on source and target emotion pairs. The top-left matrix presents the average overall results of our EmoEditor compared to the other SOTA methods. The remaining nine matrices show individual comparisons between our EmoEditor and each of the nine SOTA methods. Average preferences for our EmoEditor and other methods are indicated above each matrix. The vertical axis corresponds to the source emotion, while the horizontal axis represents the target emotion. Each cell in the matrix denotes the ratio at which our method is selected for the corresponding transition from the source to the target emotion.
    }
    \label{fig:SP_confusion_matrix}
\end{figure*}

In addition, we show the confusion matrix of different methods in Figure \ref{fig:SP_confusion_matrix}. The confusion matrix provides more detailed insights into the emotion transition effects of our EmoEditor across specific emotion categories. From the upper quadrant of the confusion matrix, it can be observed that people prefer Neural-Style-Transfer (NST) \cite{gatys2015neural} and CLIP-Styler (Csty) \cite{kwon2022clipstyler} more in the manipulation of emotions from positive to negative. This preference may arise because these methods excel in inducing negative emotions through distortions and irregular textures on the source image. In contrast, our EmoEditor achieves higher human preference scores in the lower quadrant of the confusion matrix, demonstrating our method's significant superiority in generating positive-valence images. 
Overall, for generating positive-valence images, our EmoEditor achieves the best results in generating images that evoke awe and contentment, but it is slightly less effective in generating images that evoke excitement. This may be because excitement is a more dynamic and complex emotion, requiring a more intricate combination of visual elements. 

Table \ref{tab:SP_Semo} presents the quantitative evaluation of generated images for all competitive methods. Our EmoEditor achieves the best performance in both $S_{emo}$ and ESMI. Although Ip2p scores highest on $S_{str}$, this is because it makes minimal changes to the original images in most cases, as reflected by its low $S_{emo}$ scores and visual results. In contrast, our method effectively edits the emotion-evoking regions in the source images, achieving a balance between evoking the target emotion and preserving the structure, which is supported by the highest ESMI score.

\begin{table*}[ht]
    \centering
    \setlength{\tabcolsep}{6pt} 
    \renewcommand{\arraystretch}{1.3} 
    \begin{tabular}{l|ccccccccccc}
    \hline
    Metric   & CT & NeurST & Csty & CtrlNet & SDEdit & Ip2p & AIF & EmoEdit & LMS & Ours \\ 
    \hline
    $S_{emo}$(\%)\textuparrow & 27.70 & 65.50 & 38.54 & 36.34 & 39.16 & 19.92 & 31.42 & 46.88 & 38.77 & \textbf{69.89} \\
    $S_{str}$(\%)\textuparrow & 28.96 & 29.47 & 30.11 & 29.55 & 32.65 & \textbf{34.82} & 28.00 & 31.60 & 32.03 & 33.22 \\
    \hline
    ESMI(\%)\textuparrow & 28.33 & 47.48 & 34.33 & 32.94 & 35.91 & 27.37 & 29.71 & 39.24 & 35.40 & \textbf{51.56} \\ 
    \hline
    \end{tabular}
    \vspace{-2mm}
    \caption{\textbf{Quantitative Evaluation of Generated Images for All Competitive Methods.} Best in bold. Larger (\textuparrow) is better.}
    \vspace{-2mm}
    \label{tab:SP_Semo}
\end{table*}

\begin{figure*}[ht]
    \centering
    \includegraphics[width=0.72\linewidth]{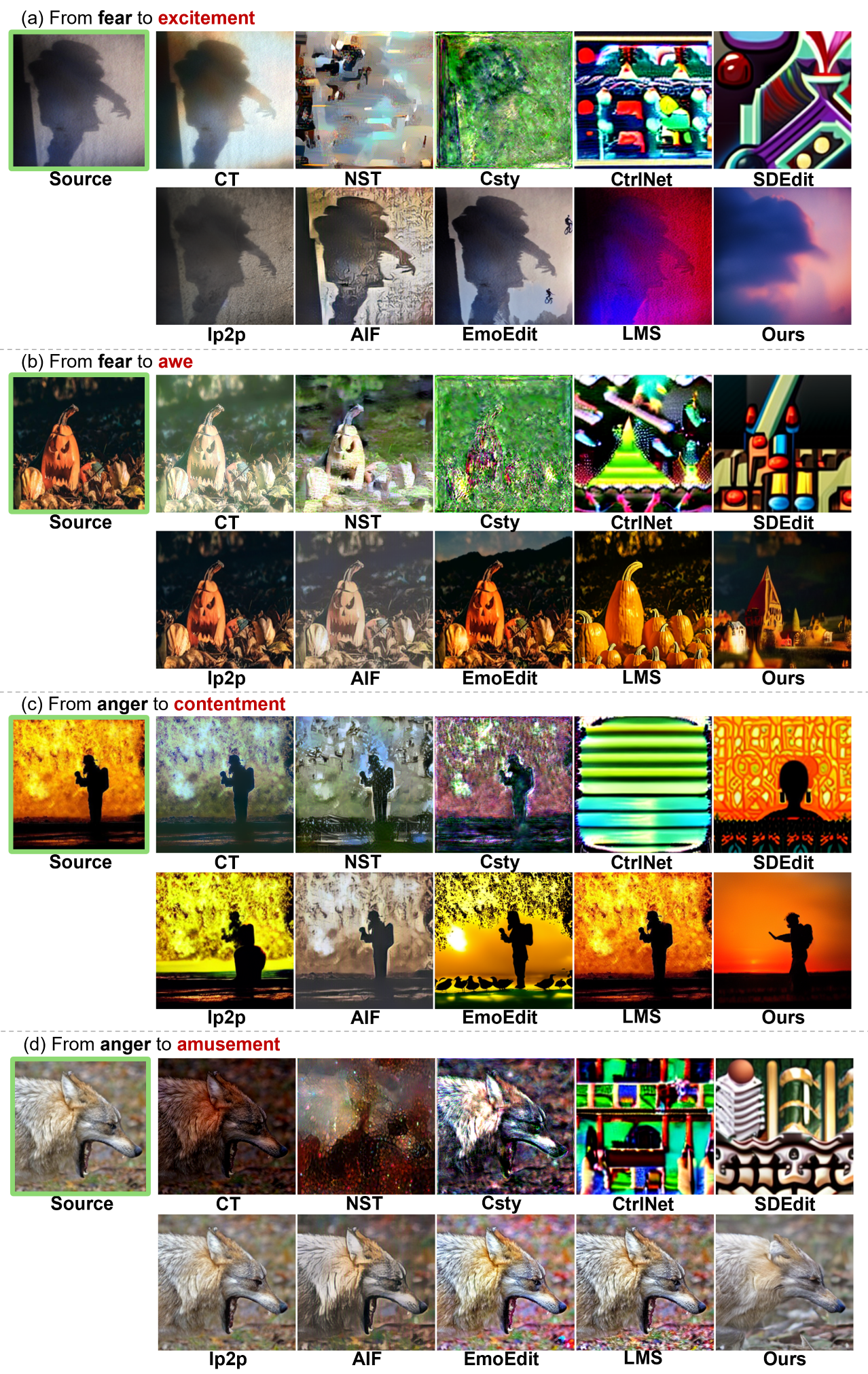}
    \caption{\textbf{Visualization of Generated Images from Different Methods (from negative to positive).} The target emotion is highlighted in red and the source image is framed in green.
    }
    \label{fig:SP_visualization}
\end{figure*}

\begin{figure*}[ht]
    \centering
    \includegraphics[width=0.72\linewidth]{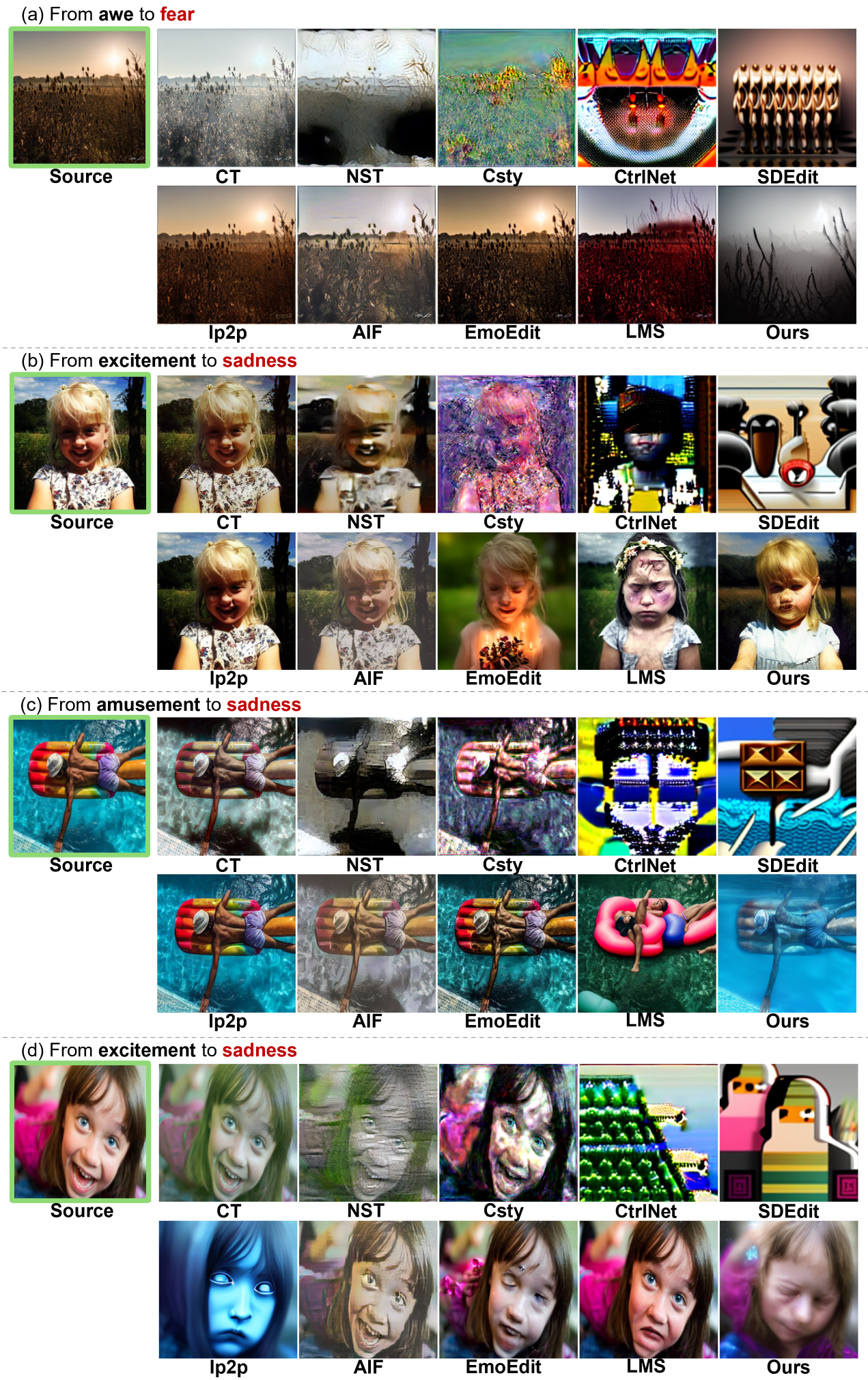}
    \caption{\textbf{Visualization of Generated Images from Different Methods (from positive to negative).} The target emotion is highlighted in red and the source image is framed in green.
    }
    \label{fig:SP_visualization_neg}
\end{figure*}

\subsection{Visualization of Emotional Image Generation across Valence}
\label{sec:SP_Visualization}

Figure \ref{fig:SP_visualization} illustrates the visualizations of images generated by all methods for the task of generating positive-valence images from negative emotions. 
Color-Transfer (CT) \cite{pitie2007automated} proves ineffective for significant emotional enhancement because it primarily replicates the color features of the reference image. For instance, in example(b), it enhances brightness but fails to evoke awe as it does not alter elements causing fear. 
Neural-Style-Transfer (NST) \cite{gatys2015neural} heavily relies on randomly selected reference images, lacking in preserving the fundamental semantic content of the source image. For example, in example(a), it generates multiple differently colored blocks but fails to evoke the emotion of excitement. 
CLIPstyler (Csty) \cite{kwon2022clipstyler}, SDEdit \cite{meng2021sdedit}, and CtrlNet \cite{zhang2023ControlNet} often introduce inexplicable textures, losing the semantics and structure of the source images.

Ip2p \cite{brooks2023instructpix2pix}, due to its limited understanding of emotions, often produces results identical to the source image, showcasing restricted image generation capabilities. 
AIF \cite{weng2023affective} heavily relies on text descriptions for understanding emotions and can only perform global edits, similar to applying a filter. As a result, it fails to achieve effective local edits to evoke the target emotion. 
EmoEdit \cite{yang2025emoedit} has difficulty accurately identifying and localizing strongly emotion-evoking regions in source images, resulting in ineffective or imprecise edits. In example (a), Ip2p, AIF, and EmoEdit fail to understand that the fear in the source image is caused by a strange shadow. As a result, they do not alter the shadow and thus cannot counteract the fear. 
Even with a series of large models in LMS, it does not consistently generate ideal outputs to trigger the target emotion. In addition, it often greatly changes the scene of the source image to achieve the goal of changing the emotion. In example (c), it fails to recognize that the large flames in the background are the primary source of the original emotion, and instead generates an image where the person remains in front of the fire, merely adjusting the flame colors to be more vibrant. Such changes are insufficient to alter the original emotional impact.

In contrast, our EmoEditor, requiring only the target emotion and source image inputs, can generate highly creative images that evoke the target emotion while striving to maintain scene structure and semantic coherence. In example(a), our EmoEditor understands that the shadow in the source image is the key factor triggering fear. Instead of altering the shadow's shape, it generates a sunset sky background, transforming the shadow into a cloud in the sky, which successfully evokes the target emotion of excitement. In example(b), it replaces a Halloween pumpkin with a castle to elicit the target emotion of awe. 
In example(c), it replaces the background flames with a red sunset in an attempt to evoke contentment. Notably, the generation process involves no manual intervention, showcasing the imagination and creativity of our model.

Additionally, Figure \ref{fig:SP_visualization_neg} shows the generation results of different methods for transitioning from positive to negative emotions. Compared to generating positive-valence images, CT and AIF perform better in generating negative-valence images. This indicates that negative emotions are more easily evoked than positive ones and can be achieved through global edits to the source image.

In addition to its creativity in generating positive-valence images, our EmoEditor also exhibits strong understanding and editing capabilities in generating negative emotions. In example(a), EmoEditor not only darkens the image’s tone to create a fearful atmosphere but also transforms the field into a desolate scene with weeds and branches, effectively evoking the emotion of fear. In example(c), it submerges the person, previously relaxing on a swim ring, beneath the pool water and alters his appearance to look frail and distressed, resembling a drowning victim, to evoke sadness.
This highlights that EmoEditor can comprehend the content of the source image, align with its original emotional state, and combine it with the target emotion to perform specific edits, addressing both global and local aspects.

\subsection{CrossTest Evaluation Results}
Our method employs an emotion predictor to guide emotional expression during image editing. However, using the same type of predictor for both generation and evaluation can introduce bias, as models may exploit specific predictor behaviors to inflate performance. To mitigate this, we split EmoSet into two disjoint subsets and trained two independent emotion predictors: one for generation and one for evaluation.

As shown in Table \ref{tab:SP_crosstest}, NST achieves the highest ESMI score. However, as illustrated in our visualization results, its outputs often feature emotion-colored overlays that obscure the original image content. These artifacts tend to inflate $S_{\text{emo}}$ because the emotion predictor is overly sensitive to low-level color cues rather than genuine emotional semantics. This exposes a key limitation of current evaluation practices and underscores the need for more semantically grounded emotion understanding.

In contrast, results from our human study (Figure \ref{fig:SP_bar}) show a clear preference for our method. Participants favored images where emotion-evoking regions were meaningfully altered, rather than simply covered with ambiguous color patches. This suggests that semantic coherence in emotional edits is more impactful and preferable than superficial visual modifications.

Our method ranks second, only 2\% below NST, while outperforming all other baselines. Unlike NST, it preserves visual semantics while achieving strong emotional alignment, demonstrating robust generalization and a more balanced trade-off between content fidelity and emotional expressiveness.

\begin{table*}[ht]
    \centering
    \setlength{\tabcolsep}{6pt} 
    \renewcommand{\arraystretch}{1.3} 
    \begin{tabular}{l|ccccccccccc}
        \hline
        ESMI(\%) \textuparrow  & CT    & NST   & Csty  & CtrlNet  & SDEdit & Ip2p  & AIF  & EmoEdit  & LMS   & Ours \\ 
        \hline
        CrossTest & 28.39 & \textbf{45.35} & 32.61 & 39.38 & 38.50 & 26.54 & 28.67 & 38.50 & 34.09 & \underline{43.95} \\
        \hline
    \end{tabular}
    \vspace{-2mm}
    \caption{
      \textbf{Quantitative Comparison of Competing Methods under CrossTest.} 
      CrossTest refers to the evaluation scheme where images are generated using an emotion predictor trained on the EmoSet subset 1 and tested by an emotion predictor trained on the EmoSet subset 2. 
      Bold indicates the best performance, while the second-best is underlined. Larger (\textuparrow) is better.
    }
    \vspace{-2mm}
    \label{tab:SP_crosstest}
\end{table*}

\subsection{Limitations}
\label{sec:SP_limitation}
\begin{figure*}[h]
    \centering
    \includegraphics[width=0.72\linewidth]{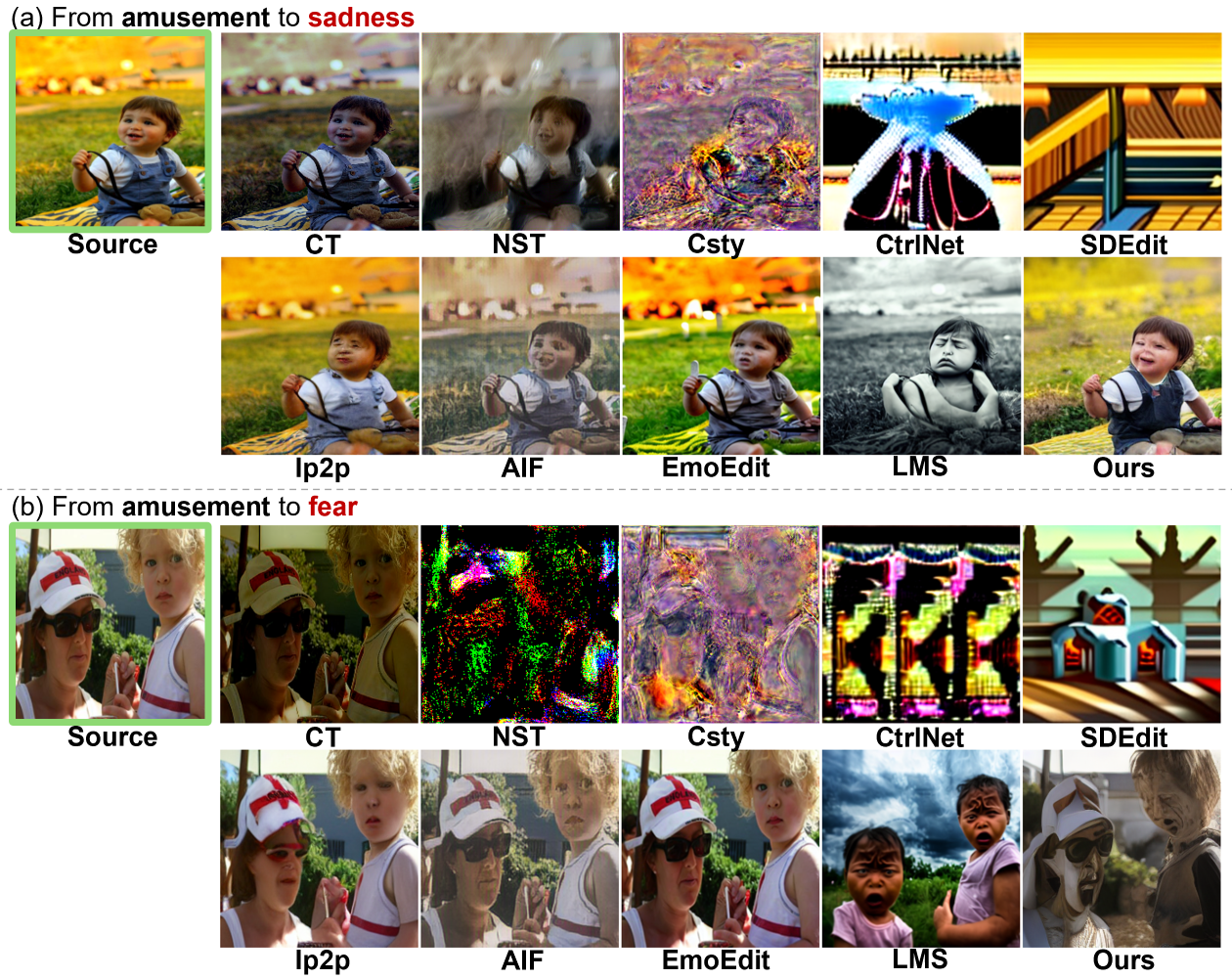}
    \caption{\textbf{Failure Case.} The target emotion is highlighted in red and the source image is framed in green.
    }
    \label{fig:fig10_failure}
\end{figure*}

While EmoEditor excels in quantitative performance and visual results, it encounters difficulties in handling fine details of individuals, especially when their faces in the images are small. For example, in Figure \ref{fig:fig10_failure}, although our EmoEditor attempts to make the boy appear sad, the boy's face seems somewhat twisted and distorted.

Moreover, although we use the 8-category Mikels model, real emotions are far more diverse and complex, and distinguishing positive emotions is often more challenging than negative ones. This is an area for future exploration.

\clearpage

\section*{Acknowledgements}
This research is supported by the National Research Foundation, Singapore under its NRFF award NRF-NRFF15-2023-0001 and Mengmi Zhang's Startup Grant from Nanyang Technological University.

{
    \small
    \bibliographystyle{ieeenat_fullname}
    \bibliography{main}
}

\end{document}